# Higher-order Segmentation via Multicuts

Jörg Hendrik Kappes[2,4], Markus Speth[4], Gerhard Reinelt[2,4], and Christoph Schnörr[1,2,4]

[1]Image and Pattern Analysis Group, Heidelberg University, Germany
[2]Heidelberg Collaboratory for Image Processing, Heidelberg University, Germany
[3]Discrete and Combinatorial Optimization Group, Heidelberg University, Germany
[4]Research Training Group 1653: Spatio/Temporal Graphical Models and Applications in Image Analysis, Heidelberg University, Germany

## Abstract

Multicuts enable to conveniently represent discrete graphical models for unsupervised and supervised image segmentation, in the case of local energy functions that exhibit symmetries. The basic Potts model and natural extensions thereof to higher-order models provide a prominent class of such objectives, that cover a broad range of segmentation problems relevant to image analysis and computer vision. We exhibit a way to systematically take into account such higher-order terms for computational inference. Furthermore, we present results of a comprehensive and competitive numerical evaluation of a variety of dedicated cutting-plane algorithms. Our approach enables the globally optimal evaluation of a significant subset of these models, without compromising runtime. Polynomially solvable relaxations are studied as well, along with advanced rounding schemes for post-processing.

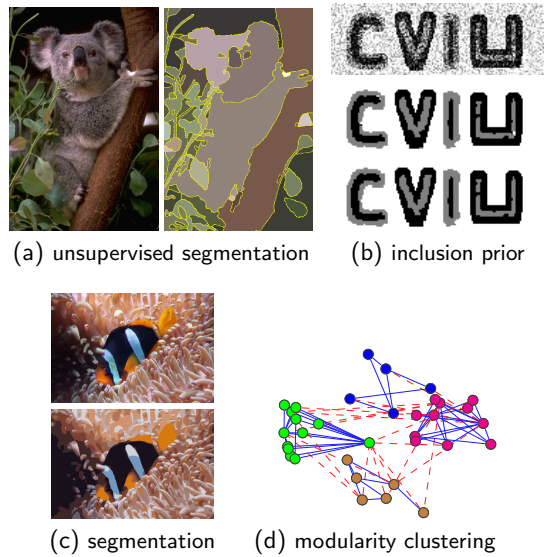

Figure 1: The presented framework covers (**a**) unsupervised and (**c**) supervised segmentation problems. In the former case, the number of components (clusters) of the partition is unknown. In the latter example, the image is partitioned (labelled) by assigning pixels to 12 predefined colors classes, taking spatial context into account. (**b**) By including higher order terms into the graphical model (bottom), segments can be enforced to include each other so as to respect topological prior knowledge. (**d**) Illustration of another example of a broad range of applications covered by the framework: graph partitioning by modularity clustering.

## 1 Introduction

### 1.1 Overview, Motivation

The segmentation problem, also known as partitioning, clustering, or grouping, is a fundamental problem of image analysis. Applications include unsupervised image partitioning [40, 4], task-specific image partitioning [41], semantic image segmentation [52, 36], and modularity clustering in network analysis [14].

Common problem representations are based on a graph $G = (V, E)$, where nodes $V$ relate to raw data on an image grid or extracted feature vectors, and edges $E$ define a neighborhood structure of the nodes. A segmentation of a graph can be represented either by

(i) assigning to each node $v \in V$ a label, or by

(ii) a multicut given by a subset of *cut edges* $E' \subseteq E$, resulting in a partition of the set of nodes $V$.

The segmentation problem is then to find a segmentation (node labeling or multicut) with minimal costs.

One commonly distinguishes *supervised* and *unsupervised* segmentation. In the former case, the number of classes represented by labels is known, together with a function measuring how likely features associated with nodes belong to each class. In the latter unsupervised case, such information is absent. This introduces ambiguities of the representation (i) since permuting the labels results in the same segmentation. Representation (ii) does not exhibit such symmetries and is therefore particularly appealing in the unsupervised case.

Accordingly, this paper focuses on the segmentation problem as a *multicut problem*, on the polyhedral representation of valid multicuts resulting in partitions of a given image [24, 16, 17], and on a computational approach to take into account the corresponding constraints efficiently.

Specifically, we consider objective functions for the segmentation problem of the form $J(x) = \sum_f \varphi_f(x_{ne(f)})$ – see Sec. 2 for details – where all higher-order terms, i.e. terms depending on more than one variable, are invariant to label permutations. For second-order models this is equivalent to Potts models that may involve negative couplings between adjacent nodes. While cutting-plane based methods [35] have shown best performance on second-order Potts models with arbitrary couplings [38], we show that terms with order higher than two can be handled by additional auxiliary variables and few additional constraints that do not interfere with the constraints defining valid multicuts. Consequently, cutting-plane meth-ods can be uniformly used for all models as we demonstrate by comprehensive numerical evaluations.

In this connection, the present paper provides a systematic comparison of different separation strategies for computer vision applications. In particular, we find that **(i)** odd-wheel inequalities do not tighten the relaxation as expected, in view of results for highly connected *non*-computer vision models [58], **ii)** integer linear programming subroutines work overall best, but **(iii)** novel extensions for separation procedures as suggested in this paper are indispensable for efficient usage.
Taking these aspects into account improves runtime by at least a factor of 2.

We also consider the supervised segmentation problem, i.e. finding an optimal multicut with at most $k$ labels, which is known as the *multiway cut problem*. Compared to the standard (I)LP representation of such problems our approach is considerably more memory efficient and able to provide globally optimal solutions for many computer vision problems in reasonable runtime [36, 37, 38]. Fig. 1 provides an overview and illustrations of the models studied in this paper.

## 1.2 Related Work

In the **unsupervised case**, the multicut polytope has recently become a focal point of research in computer vision. Major aspects of current work include closedness constraints for image segmentation [4, 7], contour completion [55], ensemble segmentation [3, 55], and the convex hull of feasible multi-cuts from the optimization point of view [35, 41, 69].

Regarding the latter viewpoint, some authors considered primal *linear program (LP)* relaxations solved by cutting-plane methods [40, 41]. Yarkony et al. [69] suggested a Lagrangian relaxation for planar graphs based on a problem decomposition into binary planar max-cut problems. Others [4, 3, 55, 35] resorted to *integer linear programs (ILPs)* as inner-loop solver within the cutting-plane formulation. While this has exponential runtime in the worst case, it may be expected to work fast in many applications. However, a comparison of these methods and variants was missing so far.

In the **supervised case**, representation (i) above prevails for the image segmentation problem [42]. Accordingly, the *marginal polytope* has become a focal point of research with respect to relaxations and approximate inference for image labeling [67, 63, 46].

Alternatively, greedy move-making algorithms like $\alpha$-expansion [13] or FastPD [48] have become established methods that are widely applied. Recently, FastPD has been generalized by [28] such that it can handle higher order models.

Methods that solve the *multiway cut problem* [16] have been considered somewhat misleadingly as computationally intractable for computer vision problems [12]. While in general this problem is known to be NP-hard [19], for few special cases, e.g., for planar graphs, exact polynomial-time algorithms are known [18, 53].

A connection of a special relaxation of the second-order multiway cut problem to variational approaches using anisotropic variants of total variation, and to the linear programming relaxation over the local polytope, has been pointed out by Osokin et al. [59] and Nieuwenhuis et al. [56]. We generalize this connection for positive and negative coupling strength in Thm. 7.1.

Recently, Kappes et al. [35] presented a cutting-plane approach to solve the multiway cut problem for various problem instances from computer vision. Globally optimal results for benchmark datasets were reported [37, 36] that compare well also in terms of runtime to state-of-the-art methods for approximate inference. However, a detailed evaluation of different separating procedures, its generalization to the higher order case as well as an analysis of the polyhedral relaxations were lacking.

Form the **modeling** point of view, models with higher-order interactions between variables have become a focus of research in the last years. Due to their enhanced expressiveness, compared to commonly used pairwise models, more complex statistics and interactions between variables can be included into models, see [68] for a more detailed discussion. The main limitation of such higher-order models was and often still is the lack of efficient inference methods, especially compared to the fast methods available for second-order models.

In order to deal with the intrinsic complexity of higher-order terms, specialized solvers have been suggested that make use of internal structures of this functions, that can be utilized for faster inference. This includes reduction techniques for higher order problems with two variables [33, 27] which have been used to generalize fusion move [51] for higher order models. For the class of (robust) $P^n$ Potts functions closed form reductions for the expansion and swap moves are known, that can be found under certain conditions in polynomial time by a min-st-cut [44, 43]. Similar techniques are used in the case of functions that depend on the *diversity* of the set of labels that its arguments (variables) take, see [26] for further details. Furthermore, graph-cut based approaches has been proposed for co-occurrence statistics [49] and label costs [21]. For message passing methods efficient update rules have been proposed for higher-order Fields-of-Experts model [50], linear constraint potentials [60] and cardinality-based potentials [66]. For linear programming relaxations sparsity of the higher-order function has be used to reduce the number of dual variables [47, 62].

In the present work we introduce a new class of higher-order functions that can be handled efficiently, namely label permutation invariant functions. These functions can be included in a much more efficient manner after a embedding of the original label space into a lower dimensional one. It is worth mentioning that the sparsity is independent on the number of labels. The class of label permutation invariant functions can be seen as a generalization of the Potts function and the reduction suggested by Kim et al. [40] as a special case, as we show in Sec. 4.4. The relation of label permutation invariant functions and $P^n$ Potts functions will be discussed in detail in Sec. 4.5.

Our label permutation invariant functions can also be used to model topology prior. As an example we show in Sec. 6.3.2 an prior that enforce that segments have to be included by not more than one other segment, see also Fig. 1(b). Delong et al. [20] suggest to solve this problem by introducing geometric interactions on the label space. While this allows a simple reduction to the second order case, the inclusion is conditioned on the labels and not on the topology of the partition, e.g. the prior of Delong et al. [20] can not restrict the label space to any sequence of including rings, because this is a pure topological prior without label dependencies, see Sec. 6.3.2 for details.

### 1.3 Contribution

1. We present a general framework for multicut problems, which includes Potts models as a special case. For the first time, we systematically compare different types of cutting-plane methods for the multicut problem in connection with computer vision applications.

2. Our framework also includes higher-order problems based on a new class of so-called *label permutation invariant (LPI) functions*. This class comprises all functions that are invariant to label permutations and thus provides a natural generalization of Potts functions.

3. We present several separation procedures and algorithmic variants that lead to significant speedups. These methods are then able to solve the problems to optimality or to provide an approximative solution in guaranteed polynomial time with bounded integrality gap.

4. We prove that these relaxed problems are equivalent to linear programs that use standard relaxations of the marginal polytope for general second-order Potts models.

5. Comprehensive numerical evaluations demonstrate the basic properties of our approach and enable us to rank the different variants.

### 1.4 Organization

We start in Sec. 2 with the problem formulation followed by introducing multicuts and corresponding problem transformations in Sec. 3. In Sec. 4 we extend the framework to higher-order models and show how corresponding higher-order terms can be taken into account in a memory-efficient way by exploiting symmetries.

We detail separation procedures for finding violated constraints in Sec. 5 and show how they can be implemented efficiently. Rounding mechanisms are discussed in Sec. 5.3. We conclude the framework with our cutting-plane method presented in Sec. 5.4 and generalize the connection to linear programs over the local polytope in Sec. 5.5.

Finally, we provide numerical evaluations for a large number of different models in Sec. 6, including second- and higher-order models in the supervised and unsupervised case, followed by concluding remarks in Sec. 7.

## 2 Problem Formulation

### 2.1 Basic Definitions

We consider discrete energy minimization problems given in terms of a *factor graph* $\mathcal{G} = (\mathcal{V}, \mathcal{F}, \mathcal{E})$, that is a bipartite graph with a set of variable nodes $\mathcal{V}$, a set of factors $\mathcal{F}$, and a corresponding relation $\mathcal{E} \subseteq \mathcal{V} \times \mathcal{F}$ associating variables to factors, cf. [45].

Variable $x_v$ assigned to node $v \in \mathcal{V}$ takes values in a discrete label-space $X_v$. We use the shorthands $X_A = \bigotimes_{v \in A} X_v$ and $x_A = (x_v)_{v \in A}$ for $A \subseteq \mathcal{V}$, in particular $X = X_{\mathcal{V}}$ and $x = x_{\mathcal{V}}$. In cases where all $X_v$ are equal we denote this label set by $L$.

Each factor $f \in \mathcal{F}$ has an associated function $\varphi_f : X_{ne(f)} \to \mathbb{R}$, where

$$ne(f) := \{ v \in \mathcal{V} \mid (v, f) \in \mathcal{E} \} \tag{1}$$

denotes the neighborhood of the factor $f$, i.e., $x_{ne(f)}$ are the variables comprising $f$. We define the *order* of a factor by the cardinality $|ne(f)|$, e.g., pairwise factors have order 2, and the order of a model by the maximal order among all factors. We denote the set of all factors of order $\mathbb{N} \ni r \geq 1$ by $\mathcal{F}_r$. The energy function of the discrete labeling problem is then given by

$$J(x) = \sum_{f \in \mathcal{F}} \varphi_f(x_{ne(f)}), \tag{2}$$

where values of the variables $x$ are also called *labelings*. We consider the problem to find a labeling with minimal energy, i.e.,

$$\hat{x} \in \arg\min_{x \in X} J(x), \tag{3}$$

for specific classes of energy functions.

By using factor graph models we take the structural property of energy functions explicitly into account. Additionally, we also consider properties of the functions $\varphi_f$. Specifically, we assume that any function with order greater than one is invariant to label permutations.

**Definition 2.1** (Label permutation invariant functions). *A function $\varphi : L^N \to \mathbb{R}$ is called invariant to label permutations if $\forall x, x' \in L^N$ with $x_i = x_j \Leftrightarrow x'_i = x'_j$ the equality $\varphi(x) = \varphi(x')$ holds.*

Many problems of interest are covered by models involving functions of this class.

Below, we will use for any predicate $\tau$ the corresponding indicator function

$$\mathbb{I}(\tau) = \begin{cases} 1, & \text{if } \tau \text{ is true,} \\ 0, & \text{otherwise.} \end{cases} \tag{4}$$

### 2.2 Supervised Case

In the supervised case we deal with energy functions (2),

$$\min_{x \in X} \sum_{f \in \mathcal{F}_1} \varphi_f(x_{ne(f)}) + \sum_{r \geq 2} \sum_{f \in \mathcal{F}_r} \varphi_f(x_{ne(f)}), \tag{P1}$$

where $\varphi_f(\cdot)$ is permutation invariant for all factors $f \in \mathcal{F}_r$, $r \geq 2$. Second-order models of this kind are known as *Potts models*, with $\mathcal{F}_r = \emptyset$ for $r > 2$ and

$$\varphi_f(x_{ne(f)}) = \beta_f \mathbb{I}(x_{ne(f)_1} \neq x_{ne(f)_2}), \qquad \forall f \in \mathcal{F}_2,$$

where $\beta_f \in \mathbb{R}$ is the *coupling constant* of factor $f$, and $ne(f)_i$, $i = 1, 2$, denotes the $i$-th neighbor of $f$. Note, that we do not restrict the models to $\beta_f \geq 0$. We focus on related higher-order models separately in Sec. 4.

### 2.3 Unsupervised Case

Contrary to the supervised problem (P1), in the unsupervised case the set of first-order factors is empty and the number of labels equals the number of variables:

$$\min_{x \in \{1, \ldots, |V|\}^{|V|}} \sum_{r \geq 2} \sum_{f \in \mathcal{F}_r} \varphi_f(x_{ne(f)}). \tag{P2}$$

In the second-order case, (P2) is known as the *pairwise correlation clustering* problem, where a set of nodes $\mathcal{V}$ has to be partitioned into clusters such that the sum of the costs of node-pairs in different clusters is minimized.

$$\min_{x \in \{1, \ldots, |V|\}^{|V|}} \sum_{f \in \mathcal{F}_2} \beta_f \mathbb{I}(x_{ne(f)_1} \neq x_{ne(f)_2}). \quad \beta_f \in \mathbb{R} \quad (P2')$$

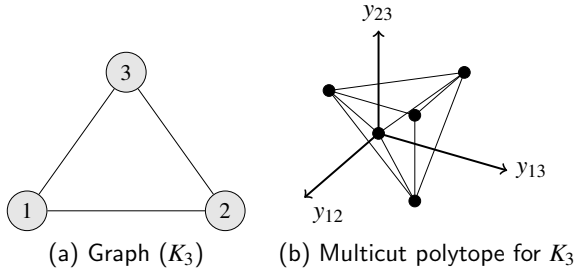

(a) Graph ($K_3$)  (b) Multicut polytope for $K_3$

Figure 2: **(a)** Illustration of the fully connected graph with three nodes $K_3$. **(b)** Illustration of the multicut polytope $MC(K_3)$, which has five vertices. Vertices of the polytope correspond to valid partitions and all other points of the polytope correspond to convex combinations of valid partitions. For large graphs the multicut polytope becomes huge and the describing system of inequalities intractable [24].

As shown in [35] for the second-order case, solving problem (P2) with solvers commonly used for problem (P1), e.g., TRWS [46], does not work, since the large state-space and label permutation invariant functions cause large sets of optimal solutions.

We study in this paper efficient methods for solving both (P1) and (P2) in the general case.

## 3 Multicuts

### 3.1 Basic Definitions

For an undirected graph $G = (V, E)$, $E \subseteq V \times V$, let $\{S_1, \ldots, S_k\}$ be a partition of $V$, i.e., $\bigcup_{i=1}^{k} S_i = V$, $S_i \cap S_j = \emptyset$, and $S_i \neq \emptyset$. We call the edge set

$$\delta(S_1, \ldots, S_k) := \{uv \in E \mid \exists i \neq j : u \in S_i \text{ and } v \in S_j\} \quad (5)$$

a *multicut* and the sets $S_i$ the *shores* of the multicut. To obtain a polyhedral representation of multicuts, we define *incidence vectors* $\chi(E') \in \{0, 1\}^{|E|}$ for each subset $E' \subseteq E$:

$$\chi_e(E') := \begin{cases} 1, & \text{if } e \in E', \\ 0, & \text{if } e \in E \setminus E'. \end{cases} \quad (6)$$

The *multicut polytope* $\text{MC}(G)$ then is given by the convex hull

$$\text{MC}(G) := \text{conv}\{\chi(\delta(S_1, \ldots, S_k)) \mid \quad (7)$$
$$\delta(S_1, \ldots, S_k) \text{ is a multicut of } G\}.$$

Fig. 2 shows an example. For further details on the geometry of this and related polytopes, we refer to [24].

The *multicut problem* is to find a multicut in a weighted undirected graph $G = (V, E, w)$, $w \in \mathbb{R}^{|E|}$, for which the sum of the weights of edges cut is minimal. Since all vertices (extreme points) of the multicut polytope correspond to multicuts, this amounts to solving the linear program

$$\min_{y \in \text{MC}(G)} \sum_{e \in E} w_e y_e. \quad \text{(P3)}$$

In order to apply linear programming techniques, we have to represent $\text{MC}(G)$ as intersection of half-spaces given by a system of affine inequalities. Since the multicut problem is NP-hard [29], we cannot expect to find a system of polynomial size. To overcome this limitation, we make use of the fact that most of these affine inequalities are not required for a given objective and use only a subset of those. The iterative construction of this subset is described in Sec. 5.

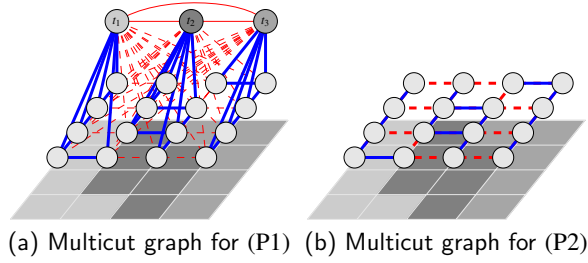

(a) Multicut graph for (P1)  (b) Multicut graph for (P2)

Figure 3: Construction of $G = (V, E, w)$ for a $4 \times 4$-grid for **(a)** the supervised case with $L = \{1, 2, 3\}$ and **(b)** the unsupervised case. Red edges are part of the multicut, i.e., they separate shores. Blue edges join nodes of the same shore of the partition.

Before discussing how problem (P3) can be solved efficiently, we show how the problems (P1) and (P2) can be transformed into problem (P3).

### 3.2 Multicuts for Second-order Models

To reformulate *problem (P2)* in the second-order case into a multicut problem we make use of the correspondence between a partition and a multicut. A given factor graph $\mathscr{G}$ defines an undirected weighted graph $G = (V, E, w)$ with $V = \mathscr{V}$, $E = \{(ne(f)_1, ne(f)_2) \mid f \in \mathscr{F}_2\}$, and $w_e = \sum_{f \in \mathscr{F}_2, ne(f) = e} \beta_f$ for all $e \in E$. Accordingly, the cost of a multicut is the sum of all $\beta_f$ over factors $f$ connecting different shores, which equals the costs of (P2) – see [16] for a formal proof and Fig. 3(b) for an illustration.

Concerning *problem (P1)* for the second-order case we assume without loss of generality that $X_v = L = \{1, \ldots, |L|\}$ for all $v \in \mathscr{V}$. Any labeling $x \in X$ defines a partition of $\mathscr{V}$. To write a second-order problem (P1) as a multicut problem (P3), we introduce additional terminal nodes $T = \{t_l \mid l \in L\} = \{t_1, \ldots, t_{|L|}\}$ and define the undirected graph $G = (V, E)$ by $V = \mathscr{V} \cup T$, $E = \{(ne(f)_1, ne(f)_2) \mid f \in \mathscr{F}_2\} \cup \{(t, v) \mid t \in T, v \in \mathscr{V}\} \cup \{(t_i, t_j) \mid 1 \leq i < j \leq |L|\}$, cf. Fig. 3(a). Thus each internal node $v \in \mathscr{V}$ is connected to all terminal nodes $t \in T$ by terminal-edges $(t, v)$.

The terminal nodes represent the $|L|$ labels $l \in L$, and label $l$ is assigned to variable $x_v$ if the terminal-edge $t_l v$ is not part of the multicut, i.e., $t_l$ and $v$ are in the same shore. Since a single label only should be assigned to each variable, $|L| - 1$ terminal-edges incident to each internal node $v$ have to be part of the multicut. This is enforced by $|\mathscr{V}|$ additional constraints given by (26) below where we take a closer look to classes of valid constraints. Edges between terminal nodes have weight 0 but are enforced to belong to different shores by additional constraints (27), which results in the so-called *multiway cut polytope*.

It remains to define the weights of terminal edges. Let $\mathbb{1}\mathbb{1}^\top$ be the matrix of all ones and $I$ be the identity matrix, both of size $|L| \times |L|$ and

$$g_v(l) = \sum_{f \in ne(v) \cap \mathscr{F}_1} \varphi_f(l), \qquad l \in L. \quad (8)$$

Then the weights $w_{t_l v}$, $l \in L$, $v \in V$, are given by

$$\begin{pmatrix} w_{t_1 v} \\ \vdots \\ w_{t_{|L|} v} \end{pmatrix} = \frac{1}{|L|-1}(\mathbb{1}\mathbb{1}^\top - I) \begin{pmatrix} g_v(1) \\ \vdots \\ g_v(|L|) \end{pmatrix}. \quad (9)$$

As before we set $w_e = \sum_{f \in \mathscr{F}_2, ne(f)=e} \beta_f$ for internal edges $e$.

# 4 Multicuts for Higher-order Label Permutation Invariant Models

We turn to higher-order models. First, we specify a class of higher-order functions which are invariant to label permutations, and show, after detailing a reduction approach, how such functions can be incorporated into a multicut framework. In its most general form, cf. Sections 4.1 and 4.3, the space complexity of representing these functions in an LP grows with the Bell number. Consequently, factors of an order more than ten are no longer tractable in this general form, cf. Tab. 1. However, for important subclasses of label permutation invariant functions efficient LP-representations exist. We will exemplary consider one such subclass of functions that can be handled even when the function comprises an order larger than several hundreds in Sec. 4.4 and show an application with factors of order 651 in Sec. 6.2.3.

## 4.1 Label Permutation Invariant (LPI) Functions

An important class of functions are label permutation invariant (LPI) functions, whose values only depend on the partitioning of the variables rather than on the labeling, as specified by Def. 2.1. They generalize Potts functions[1] (10) in a natural way.

$$f(x_1, x_2) = \begin{cases} \alpha_0 & \text{if } x_1 = x_2 \\ \alpha_1 & \text{else} \end{cases} \quad (10)$$

In other words, they can be evaluated by just knowing for all variable-pairs if their labels are identical or not. The specific label of a single variable does not matter. Any function that is evaluated on $(\mathbb{I}(x_i = x_j))_{i,j \in A}$ instead of $x_A$ itself, as e.g. the Potts function, is a label permutation invariant function. A further property is that the complexity of evaluating and storing of label permutation invariant functions does *not* grow with the number of labels.

Each possible partition of $N$ variables is uniquely represented by a binary vector over all $N(N-1)/2$ variable-pairs. But not each binary vector $\chi \in \{0,1\}^{N(N-1)/2}$ corresponds to a partition, cf. Fig. 2. The number of possible partitions is much smaller and given by the Bell numbers $B(N)$ [1]. This observation raises the issue of an efficient representation of these functions, independent of the number of labels.

Let us denote for $i = 1, \ldots, B(N)$ by $\chi_i^N \in \{0,1\}^{N(N-1)/2}$ the indicator vector of the $i$-th partition of $N$ variables. Furthermore, we define a mapping $\tau^N : L^N \to \{0,1\}^{N(N-1)/2}$ from a variable-labeling to the partition indicator by

$$\tau^N(x)_{(ij)} := \begin{cases} 1 & \text{if } x_i \neq x_j \\ 0 & \text{if } x_i = x_j \end{cases} \quad \forall 0 < i < j < N. \quad (11)$$

[1]The additional assumption that $\alpha_0$ or $\alpha_1 = 0$ can be ensured by a constant added to the function

With this we can represent any label permutation invariant function over $N = |A|$ variables parameterized by $\beta \in \mathbb{R}^{B(N)}$, where $\beta_i$ is the cost for the $i$-th partition of the sub-graph over the node-set $A$,

$$\varphi_{LPI}(x_A|\beta) = \beta_i \quad \text{if } \tau^{|A|}(x) = \chi_i^{|A|}. \quad (12)$$

As example let us consider the node-set $A = \{1,2,3\}$. There are $8 = 2^3$ different binary labelings of the edges between these nodes. Only $5 = B(3)$ of them form a valid partition and will therefore appear. We enumerate these valid partitions and assign to each a weight denoted by $\beta_i$. For a given node-labeling $x_A$ we then can calculate its binary edge-labeling $\tau^{|A|}(x)$, which corresponds one-to-one to a partition, and return the cost $\beta_i$ of this partition. Note that this function has only five parameters and does not dependent on the size of the label-space.

## 4.2 Reduction Theorem

In order to incorporate label permutation invariant functions into our multicut framework, we introduce the following reduction theorem. The basic idea of this theorem is known in the field of integer nonlinear optimization, dating back to the work of Glover et al. [31], but seemed to be unknown in other fields of research as e.g. computer vision.

**Theorem 4.1** (Reduction Theorem). *Any pseudo-Boolean function $g : \{0,1\}^M \to \mathbb{R}$ given by $g(z) = \prod_{i \in B^+} z_i \cdot \prod_{i \in B^-}(1-z_i)$, with $|B^+ \cup B^-| = M$ and $B^+ \cap B^- = \emptyset$, can be transformed into*

**(a)** *a single Boolean auxiliary variable $s \in \{0,1\}$ and two linear inequalities*

$$\min_{z \in \{0,1\}^M, s \in \{0,1\}} s \quad (13a)$$

$$\text{s.t. } Ms \leq \sum_{i \in B^+} z_i + \sum_{i \in B^-}(1-z_i) \quad (13b)$$

$$s \geq 1 - M + \sum_{i \in B^+} z_i + \sum_{i \in B^-}(1-z_i) \quad (13c)$$

*or*

**(b)** *a single auxiliary variable $s \in [0,1]$ and $M+1$ inequalities*

$$\min_{z \in \{0,1\}^M, s \in [0,1]} s \quad (14a)$$

$$\text{s.t. } s \leq z_i \quad \forall i \in B^+ \quad (14b)$$

$$s \leq (1-z_i) \quad \forall i \in B^- \quad (14c)$$

$$s \geq 1 - M + \sum_{i \in B^+} z_i + \sum_{i \in B^-}(1-z_i). \quad (14d)$$

*Proof.* The function $g(z)$ takes the value 1 if and only if $\forall i \in B^+ : z_i = 1$ and $\forall i \in B^- : z_i = 0$, and otherwise $g(z) = 0$. It remains to show that the systems of inequalities together with $s \in \{0,1\}$ or $s \in [0,1]$ restrict the feasible set such that $s = g(z)$.

Let $k$ denote the number of vanishing terms of $g(z)$

$$k = |\{i \in B^+ \mid z_i = 0\} \cup \{i \in B^- \mid z_i = 1\}|$$

then:

**(a)** Inequalities (13b) and (13c) imply

$$s \leq 1 - \frac{k}{M}, \; s \geq 1-k \quad \overset{s \in \{0,1\}}{\Rightarrow} \quad \begin{array}{l} s = 1 \quad \text{if } k = 0, \\ s = 0 \quad \text{if } k > 0. \end{array} \quad (15)$$

**(b)** Inequalities (14b)–(14d) yield

$$(14b) - (14c) \Rightarrow s \leq 0 \stackrel{s \in [0,1]}{\Rightarrow} s = 0 \quad \text{if } k > 0, \quad (16)$$

$$(14d) \Rightarrow s \geq 1 \stackrel{s \in [0,1]}{\Rightarrow} s = 1 \quad \text{if } k = 0. \quad (17)$$

□

A crucial observation is that case (b) of the reduction theorem implies integrality of $s$ if all $z_i \in \{0, 1\}$, whereas in case (a) this has to be enforced separately by $s \in \{0, 1\}$. Consequently, case (b) leads to tighter relaxations by only enforcing $s \in [0, 1]$.

While reduction (b) thus seems to be preferable, due to a lower number of constraints, reduction (a) can be nevertheless appealing for some (I)LP-solver techniques, e.g., dual simplex. The reason is that sometimes a single constraint (13b) is a less tight but a sufficient relaxation compared to several small constraints (14b) and (14c)[2]. In our experiments, we therefore use all $M+2$ constraints (13b),(13c) and (14b), (14c) (note that (13c) equals (14d)), and let the solver choose the active constraint set.

### 4.3 Reduction for Label Permutation Invariant Functions

In order to apply Thm. 4.1 to a label permutation invariant function (12) of order $N = |A|$ we rewrite it as a pseudo-Boolean function

$$\varphi_{LPI}(x_A | \beta) = \sum_{i=1}^{B(N)} \beta_i \cdot \underbrace{\prod_{j=1}^{\frac{N(N-1)}{2}} l\left([\chi_i^N]_j, [\tau^N(x)]_j\right)}_{g_i(\tau^N(x))} \quad (18)$$

with

$$l(b_1, b_2) = \begin{cases} 1 - b_2, & \text{if } b_1 = 0, \\ b_2, & \text{if } b_1 = 1. \end{cases} \quad (19)$$

We apply the reduction theorem to each of the $B(N)$ binary functions $g_i(z)$, $z = \tau^N(x)$. Consequently, a function $\varphi_{LPI}(x_A | \beta)$ of order $N$ requires $B(N)$ auxiliary variables. These auxiliary variables are connected to the node-variables via the Boolean expressions $l(\cdot, \cdot)$ in (19) and correspond to the edge-variables $y$ used in the multicut representation (P3). By this, we also get rid of difficulties caused by ambiguities of the node-label representation of a partition.

If an expression $l(\cdot, \cdot)$ has no corresponding edge $e$ in $G$, we add this edge to $G$ with weight zero.

Summing up, to include an label permutation invariant factor of order $N$ into our multicut framework, we require at most $N(N-1)/2$ edge variables, $B(N)$ auxiliary variables, and $B(N) \cdot (N(N-1)/2 + 2)$ linear inequalities. These numbers are upper bounds, of course. In many cases more compact representations are obtained.

We observed in numerous experiments that additionally enforcing that all auxiliary variables corresponding to a higher-order term sum up to 1 significantly speeds up optimization. This entails to complement a single equality constraint for each higher-order term.

Fig. 4(a) illustrates an example of a factor of order three. The reduction requires $B(3) = 5$ auxiliary variables corresponding to possible partitions and, correspondingly, they are denoted by 000, ..., 111 in the figure. Constraints generated by the reduction theorem relate these auxiliary variables to the original higher-order problem. A single additional edge shown dotted in Fig. 4(a), has to be added to the graph $G$ in this example.

[2] Consider the case $B^+ = \{1, 2\}$, $z_1 = 0.1$ and $z_1 = 0.3$. Eqns. (13b) and (14b) give $s \leq 2$ and $s \leq 0.1$, respectively.

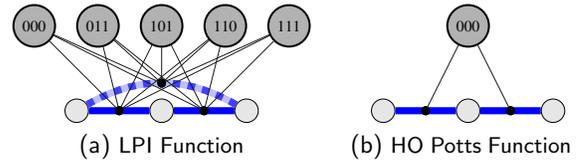

(a) LPI Function  (b) HO Potts Function

Figure 4: Higher-order label permutation invariant functions are dealt with by problem reduction and additional binary auxiliary variables (Sec. 4). Corresponding constraints (black lines) enable to represent exactly the original higher-order problem. Panel **(a)** shows an example of a generalized Potts function of order three. Panel **(b)** shows an example of a Potts function of order three.

### 4.4 Higher-Order Potts Functions

A subclass of label permutation invariant functions that can be handled more efficiently, are functions taking the value $\alpha_0$ if all variables $x_A$ with $A \subseteq V$ have the same label (are in the same shore) and $\alpha_1$ otherwise. We call such functions *higher-order (HO) Potts functions* since they constitute the simplest generalization of (second-order) Potts functions to the higher-order case. Such functions are general enough to model the costs of a hyper-graph partitioning, which was used in [40, 34]. The cost for a hyper-edge is included in the overall cost function if the hyper-edge connects at least two shores:

$$\varphi_{HOP}(x_A | \alpha) = \begin{cases} \alpha_0, & \text{if } \forall i, j \in A : x_i = x_j, \\ \alpha_1, & \text{else.} \end{cases} \quad (20)$$

We can reformulate such functions in a pseudo-Boolean form:

$$\varphi_{HOP}(x_A | \alpha) = \alpha_1 + (\alpha_0 - \alpha_1) \prod_{e \in E_A} (1 - y_e) \quad (21)$$

where $E_A$ is a subset of the edges of $G$ that spans $A$. If $G_A = (A, E \cap (A \times A))$ is disconnected we have to add some edges with weight 0. We point out our empirical observation that using a spanning graph that includes all edges of $G_A$, instead of an arbitrary spanning-tree, leads to shorter runtimes.

As before, we apply the reduction theorem to add a higher-order Potts function as part of a model at hand. This only requires a single auxiliary variable. Fig. 4(b) provides a sketch for a function of order three.

Tab. 1 shows the number auxiliary variables, size of auxiliary factors and number of additional constraints needed by the different reductions to add a HO-Potts Function with $L$ labels and order $N$. Including a higher-order Potts function as a factor-table requires no additional variables but a factor with $L^N$ entries. $L \cdot N$ constraints are only required when it is reformulated into an LP. When we use the generic reduction for LPI functions we make no use of all symmetries and need $B(N)$ auxiliary variables and factors, together with $B(N) \cdot (N + 2) + 1$ constraints. While no additional constraints are needed for including this into an LP, the Bell number $B(N)$ makes this intractable for larger $N$'s. When using HO-Potts functions we need only one single auxiliary variable together with $N+2$ constraints. This is much smaller than all other alternatives and does not depend on the number of labels.

| Reduction Method | variables | factors | size | constraints |
|---|---|---|---|---|
| higher order function | 0 | | $L^N$ | $0^*$ |
| LPI function | $B(N)$ | $B(N)$ | | $B(N) \cdot (N+2) + 1$ |
| HO-Potts function | 1 | 1 | | $N+2$ |
| $P^n$-Potts function | $L$ | | $N \cdot L + L$ | $0^*$ |

Table 1: **Comparison of reductions for HO-Potts:** Table above shows the resources required by different methods to reduce a higher order Potts factor of order $(N)$ with $L$ labels per variable. The naive method would just add a higher-order factor. The general reduction for LPI functions does not depend on the number of labels, but grows with the Bell number of the order $(B(N))$ and is therefore limited. The specialized reduction HO-Potts, grows linear with the order and is independent on the number of labels. Alternatively, $P^n$-Potts are applicable for this type of functions. However, this reduction depends on the number of labels (L).

### 4.5 $P^n$ Potts Functions

Another generalization of Potts functions for higher order called $P^N$ Potts (22) has been suggested by Kohli et al. [44].

$$f(x_A) = \begin{cases} \gamma_k & \text{if } x_i = k \ \forall i \in A \\ \bar{\gamma} & \text{otherwise} \end{cases}, \quad \gamma_k \leq \bar{\gamma}, \forall k \quad (22)$$

The assumption that $\gamma_k \leq \bar{\gamma} \ \forall k$ ensures sub-modular auxiliary problems [44]. From a modeling point of view this is not necessary.

If all $\gamma_k$ are equal and $|A| = 2$, then (22) is equivalent to Potts function (10). When we enforce $\gamma_k \leq \bar{\gamma} \ \forall k$ it is equivalent to Potts function with positive coupling, denoted as Potts+ in Fig. 5.

If $\gamma_k$ vary with $k$ the function is no longer invariant to label permutation.However, $P^n$-Potts functions are not powerful enough to model all label permutation invariant functions. For example $P^n$-Potts functions assign to $(1,1,1,2,3)$ and $(1,1,1,2,2)$ always the same energy, but those are different partitions into 3 and 2 clusters, respectively. The same hold for robust $P^n$-Potts functions (23) [44]

$$f(x_A) = \min\left\{\min_k\left\{\gamma_k + \sum_{i \in A} \frac{\bar{\gamma} - \gamma_k}{Q} \mathbb{I}(x_i \neq k)\right\}, \bar{\gamma}\right\},$$

$$\text{with } \gamma_k \leq \bar{\gamma} \ \forall k \text{ and } Q < \frac{|A|}{2} \quad (23)$$

which cover a large subclass functions but still only a small subset of LPI functions.

Fig.5 illustrate the relation between $P^N$ Potts and LPI functions. Even if we would not enforce that $\gamma_k \leq \bar{\gamma} \ \forall k$ then $P^N$-Potts would not include all LPI functions. The intersection between $P^N$-Potts and label LPI-functions are HO-Potts functions (12). A reduction of a HO-Potts functions as $P^n$-Potts function needs $L$ auxiliary variables, $L$ factors of size $N$ and $L$ of size 1. Constraints are only required when it is reformulated into an LP. Especially for large $L$ this is much more expensive than the reduction introduced in Sec. 4.4. As for Potts functions the framework suggested in [44] does only cover cases where $\gamma_k \leq \bar{\gamma}$.

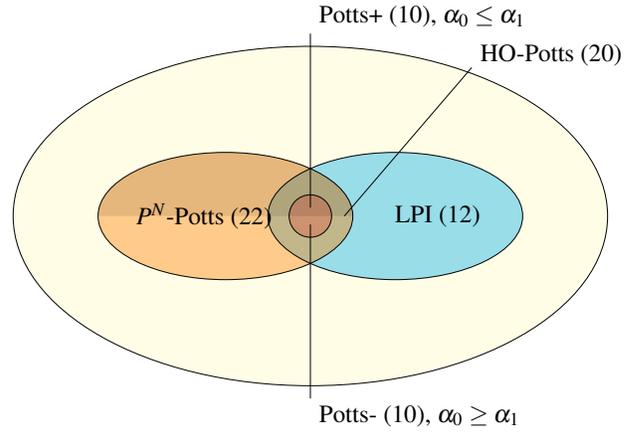

Figure 5: **Classes of Higher-Order Functions:** With in the class of higher-order functions $P^N$-Potts and LPI functions are two major subclasses. The original definition of $P^N$ Potts functions (22) include only the none-shaded region. If we remove the constrains on $\bar{\gamma}$ we get a more general class. The intersection of $P^N$ Potts and LPI functions are HO-Potts functions, cf. Sec. 4.4, which includes Potts functions with positive (Potts+) and negative (Potts-) coupling strength as special case.

## 5 Cutting-Plane Approach and Separation Procedures

### 5.1 Approach

Determining a multicut with minimal costs is NP-hard in general [29]. However, if given data induce some structure then it is plausible to expect such problems to be easier solvable in practice, than problems without any structure.

We use a cutting-plane approach to iteratively tighten an outer relaxation of the form

$$\arg\min_{y \in Y} \sum_{e \in E} w_e y_e. \quad (24)$$

Here, $Y \supseteq \text{MC}(G)$ is superset of the multicut polytope $\text{MC}(G)$ (cf. (P3)) or $\{0,1\}^{|E|} \supset Y \supseteq \text{MC}(G) \cap \{0,1\}^{|E|}$ in the integer case. In each step we solve a problem relaxation in terms of a linear or integer linear program, detect violated constraints from a pre-specified finite list (cf. Sec. 5.2) and augment the constraint system accordingly. This *separation procedure* is repeated until no more violated constraints are found.

After each iteration we obtain a lower bound as the solution of the (I)LP and an upper bound by mapping the obtained solution to the set of feasible points (rounding, cf. Sec. 5.3).

### 5.2 Relaxation, Constraints

#### 5.2.1 Initial Constraints

We start with a polytope that enforces any edge-variable $y_e$ to be lower and upper bounded by 0 and 1, respectively,

$$y_e \in [0,1], \quad \forall e \in E \quad (25)$$

In presence of terminal nodes, we additionally enforce for each non-terminal node $v \in V \setminus T$ that exactly one incident

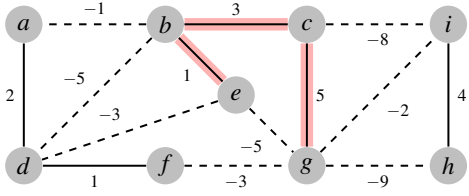

Figure 6: Edges labelings have to be constrained in order to be consistent. The cut edges are shown dashed in the figure. This edge labeling is inconsistent since it does not respect the transitivity of the corresponding relation: being in the same segment. For example, the red path implies by transitivity that $e$ and $g$ are in the same segment, in conflict to the edge-label of the edge $eg$.

edge is not cut, i.e.,

$$\sum_{t \in T} y_{tv} = |T| - 1, \qquad \text{if } T \neq \emptyset, \forall v \in V \setminus T. \quad (26)$$

Furthermore, we add the compulsory constraints

$$y_{tt'} = 1, \qquad \forall t, t' \in T, t \neq t', \quad (27)$$

forcing different terminal nodes to belong to different shores.

#### 5.2.2 Integer Constraints

A more restrictive alternative to (25) are the integer constraints

$$y_e \in \{0, 1\}, \qquad \forall e \in E. \quad (28)$$

Note that not every vector $y \in \{0,1\}^{|E|}$ belongs to the multicut polytope. Hence, even enforcing Boolean variable values may lead to inconsistent edge-labelings, cf. Fig. 6, which will be discussed in more detail in the following subsection. In general, using constraints (28) renders inference problems more difficult. On the other hand, finding violated constraints can be much simpler for Boolean-valued variables than for less tight non-Boolean relaxations. This may well compensate the additional costs for solving an ILP instead of an LP[3].

#### 5.2.3 Cycle Constraints

The problem of inconsistent edge-labelings has been considered in the literature, either motivated by closing contours [55, 4] or as tightening the multicut polytope relaxation via cycle constraints [17, 58, 40, 35]. In both cases inconsistent cycles are detected. If integer constraints are enforced an inconsistent cycle is a cycle that contains exactly a single cut edge, which obviously violates transitivity. This can be generalized to the relaxed non-Boolean case $y_e \in [0, 1]$ [17].

A system of *cycle inequalities* that necessarily has to be satisfied by consistent labelings, is given by

$$\sum_{e \in P} y_e \geq y_{uv} \qquad \forall uv \in E, P \in \text{Path}(u,v) \subseteq E. \quad (29)$$

It is well known [17] that if and only if the cycle $\{uv\} \cup P$ is chordless, then the constraint is facet-defining for the underlying polytope or, speaking less technically, "effective" for enforcing labeling consistency.

While for fully connected graphs, (29) can be represented by a polynomial number of triangle constraints [17, 32, 14],

---

[3]Note, sometimes solving the ILP is even faster than the LP.

the separation procedure reduces to a sequence of shortest path problems in the general case [17]. Given $y$, the naive approach searches for each edge $uv \in E$ the shortest path from $u$ to $v$ in the weighted graph $G_y = (V, E, y)$. If this path is shorter than $y_{uv}$, then it represents the most violated constraint of the form (29) for $uv$. Using a basic implementation of Dijkstra (as we do) the cost for one search is $O(|V|^2)$. The cost can be reduced to $O(|E| + |V| \log |V|)$ by using Fibonacci heaps.

To reduce the number of shortest path searches we exploit the following three ideas:

**Efficient Bounds on the Shortest Path (B):** Instead of searching for each edge $uv \in E$ the shortest path from $u$ to $v$ in a positive weighted graph $G = (V, E, y)$, we can calculate a lower bound on the path length for all $uv \in E$ in $O(|E| + |V|)$. To this end, we determine the connected components in the graph $G' = (V, \{e \in E \mid y_e < \gamma\})$. If two nodes $u, v \in V$ are not in the same connected component, the shortest path from $u$ to $v$ is greater than or equal to $\gamma$. Choosing $\gamma = 1$ yields a preprocessing procedure that enables to omit many shortest path searches. Furthermore, if the edge between two nodes has weight 0, this is obviously the shortest path since all edge-weights $y_e$ are non-negative.

**Shortest Path in Binary Weighted Graph (I):** If the edge weights are either 0 or 1, then simple breadth-first search can be applied instead of the Dijkstra algorithm. The computational effort can be further reduced, as before but without additional costs, by restricting the search to the graph $G_0 = (V, \{e \in E \mid y_e = 0\})$. Since any path including an edge with weight 1 cannot be shorter than the edge between the two nodes which is 0 or 1.

**Finding Chordless Shortest Paths / Facet-Defining Constraints (F):** A path between the two nodes forming an edge is called *chordless* if the cycle consisting of the path and the edge has no chord. Shortest path search can be easily extended so as to determine the shortest chordless paths: Every node except for the end-node is not updated by the Dijkstra algorithm if the path from this node to the starting node is chordal. This increases the costs by a factor bounded by $|V|$. In view of cycle constraints, the corresponding constraints are facet-defining.

Our experiments, discussed by Fig. 7 and in Sec. 6, spot that joint application of bounding procedures, facet-defining constraints (chordless paths) and dedicated search methods for binary weighted graphs, leads to better runtimes in nearly all cases.

#### 5.2.4 Terminal Cycle Constraints

In the supervised case we can further reduce the costs for shortest path searches based on the following lemma.

**Lemma 5.1.** *In the presence of terminals there exists no cycle $C$ with more than three nodes that is chordless and contains a terminal node.*

*Proof.* Let $C$ be a cycle with more than three nodes that contains a terminal node $t$, and select an edge $uv$ in $C$ with $u, v \neq t$. The $tu, tv \in E$ by definition, hence the cycle is chordal. □

As a result of Lemma 5.1, we ignore all cycle constraint of a length greater than 3 that includes a terminal node. All facet-defining cycle constraints that include a terminal node

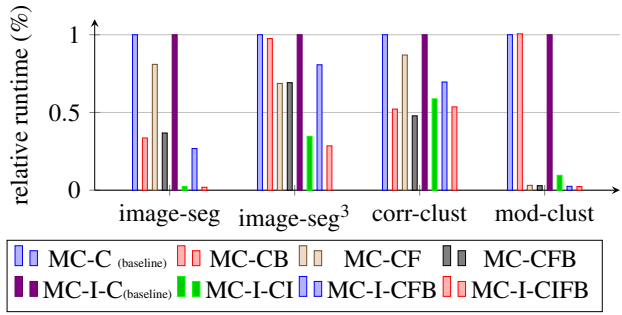

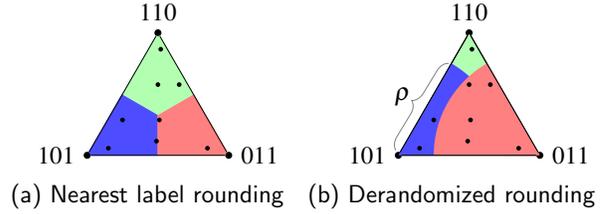

(a) Nearest label rounding  (b) Derandomized rounding

Figure 8: Illustration of the two rounding schemes for the multiway cut problem for the vector $(y_{tv})_{t \in T}$. *Nearest label rounding* (a) assigns each point in the simplex to the nearest vertex. Fig. (b) shows exemplarily one iteration of *derandomized rounding* for $\rho = 0.75$.

Figure 7: Comparison of the proposed extensions (marked by the postfixes B, I, and F) on the runtime. For the relaxed case (MC-C∗) (first four bars) we observe that bounding clearly improves runtimes for image-based data, this is not true for third-order image segmentation and modularity clustering. Using only facet-defining constraints decreases the runtime for all four datasets, most significantly for modularity clustering. If we enforce integrality (MC-I-C∗) during the cutting-plane procedure (last four bars), the use of specialized search methods (CI) reduces the runtime significantly.

are then given by

$$y_{tu} + y_{tv} \geq y_{uv}, \qquad \forall uv \in E, t \in T, \quad (30)$$
$$y_{tu} + y_{uv} \geq y_{tv}, \qquad \forall uv \in E, t \in T, \quad (31)$$
$$y_{tv} + y_{uv} \geq y_{tu}, \qquad \forall uv \in E, t \in T, \quad (32)$$

together with (27). As a consequence we only have to search for general cycle constraints on the graph without terminal nodes, that has $|T| \cdot |V|$ fewer edges!

### 5.2.5 Multi Terminal Constraints

Calinescu et al. [15] suggested another class of non-facet-defining linear inequalities for the supervised case that further tightens the outer polytope relaxation:

$$y_{uv} \geq \sum_{t \in S}(y_{tu} - y_{tv}), \qquad \forall uv \in E, S \subseteq T. \quad (33)$$

Intuitively, these constraints enforce each non-terminal edge to be at least as "much cut" as all its terminal edge-pairs indicate. Since $\sum_{t \in T}(y_{tu} - y_{tv}) = 0$, we only consider differences in the direction $u \to v$. An alternative representation of (33) exploiting symmetry is

$$y_{uv} \geq \sum_{t \in T} \frac{1}{2}|y_{tu} - y_{tv}|. \quad (34)$$

In order to see why multi terminal constraints are useful, let us consider a tiny toy example of a model with two variables and four labels. Overall, the multiway cut polytope has eight terminal edges $(t,1)_{t \in T}, (t,2)_{t \in T}$ and a single edge $(1,2)$ between the two nodes. We inspect few values of $y$ and check if (33) is implied by (30)–(32) or not.

| $(y_{t,1})_{t \in T}$ | $(y_{t,2})_{t \in T}$ | | (30)–(32) | (33) |
|---|---|---|---|---|
| $(1,1,1,0)$ | $(1,1,0,1)$ | $\Rightarrow$ | $1 \leq y_{12} \leq 1$ | $1 \leq y_{12}$ |
| $(1,1,\frac{1}{2},\frac{1}{2})$ | $(1,1,\frac{1}{2},\frac{1}{2})$ | $\Rightarrow$ | $0 \leq y_{12} \leq 1$ | $0 \leq y_{12}$ |
| $(\frac{1}{2},\frac{1}{2},1,1)$ | $(1,1,\frac{1}{2},\frac{1}{2})$ | $\Rightarrow$ | $\frac{1}{2} \leq y_{12} \leq \frac{3}{2}$ | $\mathbf{1} \leq y_{12}$ |
| $(1,\frac{2}{10},\frac{3}{10},\frac{5}{10})$ | $(1,\frac{1}{10},\frac{2}{10},\frac{7}{10})$ | $\Rightarrow$ | $\frac{2}{10} \leq y_{12} \leq \frac{3}{10}$ | $\frac{2}{10} \leq y_{12}$ |

In the third example (row) above, multi terminal constraints tighten the relaxation. It can be shown that these constraints may tighten the relaxation only if at least four terminal nodes are present.

### 5.2.6 Odd-Wheel Constraints

While cycle constraints are only sufficient to obtain optimal solutions if integer constraints are enforced, we may tighten the relaxation in the case $y_e \in [0,1]$ by adding more complex constraints.

One such a class of constraints for which the separation procedure can be carried out efficiently, are *odd-wheel constraints*. A *wheel* $W = (V_W, E_W)$ is a graph with a selected center node $c \in V_W$. All other nodes are connected with the center, and the remaining edges build a cycle containing all nodes in $V_W \setminus \{c\}$. An *odd-wheel* is a wheel with an odd number of non-center nodes. The *odd-wheel constraints* are given by

$$\sum_{uv \in E_W, u,v \neq c} w_{uv} - \sum_{v \in V_W \setminus \{c\}} w_{cv} \leq \left\lfloor \frac{||V_W| - 1|}{2} \right\rfloor \quad (35)$$

for all odd-wheels $W = (V_W, E_W)$.

Deza et al. [25] proved that odd-wheel constraints are facet-defining for $||V_W| - 1| \geq 3$. As described in detail by Deza and Laurent [23] and Nowozin [57], the search for violated odd-wheel constraints can be reduced to a polynomial number of shortest path searches, if the current solution does not violate any cycle constraints.

In our experiments, we found that with increasing sparsity, odd-wheel constraints tighten the relaxation less. This is intuitively plausible since in densely connected graphs significantly more odd-wheels exist that could be violated. Since the overall gain was not better than with the previously proposed methods, we did not spend time to search for heuristics to speed up computation, as we did for the cycle inequalities.

## 5.3 Rounding Fractional Solutions

Relaxations of the integer-valued multicut problem yield solutions that may be fractional and therefore infeasible. The objective value then is a lower bound of the optimal value. The procedure to map an infeasible solution to the feasible set is called *rounding*. Furthermore, for the resulting multicut, a corresponding node-labeling has to be determined.

### 5.3.1 Supervised Case

*In the presence of terminal nodes*, we assign to each node-variable the label of the terminal node to which it is connected by means of $y_{tv} = 0$ in the integer-valued case. This idea extends to the general case by assigning to node $v$ the label $l$ with the lowest edge-value $y_{t_l v}$, i.e., the nearest corner in the corresponding simplex, cf. Fig. 8:

$$x_v = \arg\min_{t \in T} y_{tv} \qquad \forall v \in V \setminus T. \qquad (36)$$

This heuristic *nearest label rounding* method has two drawbacks, however. Firstly, it does not provide any performance guarantee. Secondly, nearby nodes that favor two or more labels nearly equally might be randomly assigned to different labels due to numerical inaccuracy. This is particularly problematic in case of positive coupling strengths where homogeneously labeled regions are preferred.

Contrary to this local procedure, Calinescu et al. [15] suggested a randomized rounding procedure that provides optimality bounds for Potts models with positive coupling strengths. Given a threshold $\rho \in [0, 1]$, they iterate over all labels in a fixed order and assign label $l$ to node $v$ if $y_{t_l v} \leq \rho$ and no label was assigned to $v$ before. In case no label was assigned to node $v$ in the end, then the last label with respect to the ordering of the labels is assigned to $v$. This rounding procedure is sketched by Fig. 8(b).

A *randomized rounding* procedure would apply this for all $\rho \in [0, 1]$ and select the labeling with the lowest energy. Since $[0, 1]$ is uncountable, Călinescu et al. suggested a derandomized version. This is based on the observation that we only have to consider $|V \setminus T| \cdot |T|$ different threshold parameters, namely the values of the terminal edge variables $y_{tv}$. Since this set can still be quite large, we also consider a heuristic approximation that we call *pseudo-derandomized rounding*, using a small number of equidistant thresholds, in practice: $0, 0.01, 0.02, \ldots, 0.99, 1$.

Concerning tightness of the relaxation, Calinescu et al. [15] pointed out that the integrality ratio of the relaxed LP for the second-order multiway cut problem with positive coupling strengths, exploiting cycle, terminal and multi-terminal constraints, is $\frac{3}{2} - \frac{1}{k}$. This is superior to the $\alpha$-expansion algorithm [13] and the work of Dahlhaus et al. [18], which guarantees only a ratio of $2 - \frac{2}{k}$. It is not known if these results can be extended for higher-order label permutation invariant functions.

Empirically, we observe for these types of models that derandomized rounding and pseudo-derandomized rounding usually lead to results that are slightly better than when using nearest label rounding. While pseudo-derandomization does empirically not give results worse than original derandomization, it is much faster, but does not come along with theoretical guarantees. Fig. 9 shows results for two instances taken from [36]. While for the synthetic instances rounding matters, for real world examples the differences are negligible.

### 5.3.2 Unsupervised Case

*In absence of terminal nodes*, we compute in the integer-valued case the connected components of $G_0 = (V, \{e \in E \mid y_e = 0\})$, enumerate them by $\#CC_{G_0}$, and assign to each node-variable as label the number of its connected component

$$x_v = \#CC_{G_0}(v), \qquad \forall v \in V. \qquad (37)$$

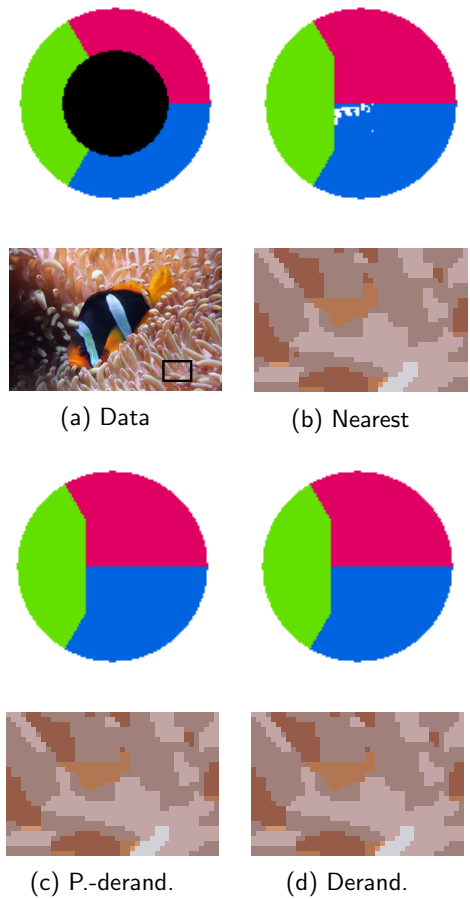

(a) Data  (b) Nearest

(c) P.-derand.  (d) Derand.

Figure 9: Illustration of the rounding results (nearest label, pseudo-derandomized and derandomized) after solving the LP relaxation with terminal, multi-terminal, and cycle inequalities for the instances inpainting and clownfish from [36]. Derandomized and pseudo-derandomized rounding gives similar results. Simple rounding to the nearest label can give inferior results (top row). But for real applications differences of the labelings are marginal (last row).

It is easy to see that the labeling-costs $J(x)$ (2) are greater than or equal to the multicut costs $\langle w, y \rangle$ and equal if $y$ is a valid multicut.

If $y$ is not integral we first have to map $y$ to a vertex of the multicut polytope. To this end, we determine the connected components of $G_{\leq \kappa} = (V, \{e \in E \mid y_e \leq \kappa\})$ and define the feasible projection $\hat{y}$ by

$$\hat{y}_{uv} = \begin{cases} 0, & \text{if } \#CC_{G_{\leq \kappa}}(u) = \#CC_{G_{\leq \kappa}}(v), \\ 1, & \text{else.} \end{cases} \qquad (38)$$

The labeling then is given by

$$x_v = \#CC_{G_{\leq \kappa}}(v), \qquad \forall v \in V. \qquad (39)$$

Since the connected component procedure tends to remove dangling edges, it seems to be reasonable to select $\kappa$ smaller than 0.5. This was empirically confirmed by our experiments. Fig. 10 shows the relative error of the rounded solutions after enforcing cycle constraints for different problem-classes with various values of $\kappa$.

It is worth to mention that the multicut problem is APX-hard [8, 22] and that a rounding procedure with a worst case integrality gap of the linear program of $\Omega(\log(|V|))$ exists. While it seems that the corresponding proof can be extended to higher-order multicuts, an integrality gap guarantee of $\log(|V|)$ is not enough in real applications. This is why we did not further investigate this issue in the present work.

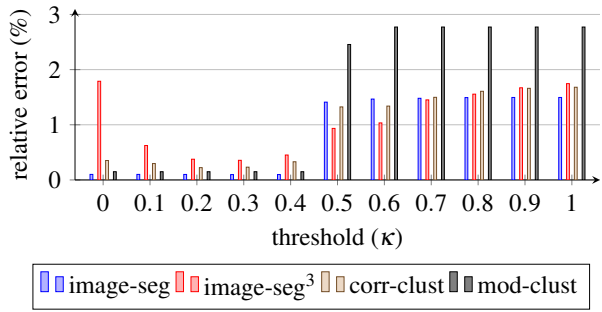

Figure 10: Illustration of the impact of the choice of $\kappa$ on the distance of the energy of the integer solution obtained by rounding to the optimal value. For modularity clustering (mod-clust) and third-order image segmentation (image-seg$^3$) we scaled the bars by a factor of 0.1. The results show that one should choose $\kappa < 0.5$. Empirically the optimal value lies in $[0.2, 0.3]$ but also 0 (more precisely $10^{-8}$) gives nearly similar results.

### 5.4 Multicut Cutting-Plane Algorithm

Alg. 1 provides a compact description of our complete multicut approach, summarizing the present section. In addition to the specification of the objective function in terms of a factor graph model $\mathscr{G}$, we expect a proper[4] list of separation procedure sets $\mathbb{S}$ as input parameters. For example, $\mathbb{S}_1$ could represent simple cycle constraints separation, $\mathbb{S}_2$ integrality constraints, and $\mathbb{S}_3$ cycle constraints separation specialized to integer solutions.

As specified by Alg. 1, we construct the weighted undirected graph $G$, introduce auxiliary variables for higher-order factors (as detailed in previous sections), and initialize the constraint set $\mathscr{C}$ by a simple outer relaxation of the feasible set.

For each separation procedure set in the list $\mathbb{S}$, we apply all separation procedures in $\mathbb{S}_i$ to find violated constraints and add these to $\mathscr{C}$ until no more are found. Then we proceed with the next set $\mathbb{S}_{i+1}$.

The (integer) linear program in line 6 is solved by CPLEX 12.2, a standard off-the-shelf LP-solver. Finally, we compute an optimal node-labeling $x \in X$ from the multicut solution $y$.

The implementation of Alg. 1 turned out to be involved, due to several subtle pitfalls necessitating some care. We therefore made our code publicly available[5]. Furthermore, when solving the (I)LP one should not expect that the solution is feasible. Sometimes we observe negative values of $y_e$ and therefore project solutions always to $[0,1]^{|E|}$. Also Boolean constraints were sometimes slightly violated. Most importantly, due to numerical reasons, constraints should only be added if they are significantly violated, i.e., the constraint

---
[4]A list of separation procedures is called proper if the separation procedures that are included once are also included when proceeding further down the list. For proper lists the obtained relaxation is well-defined. All lists used in our experiments are proper.
[5]https://github.com/opengm/opengm

---

**Algorithm 1** Multicut-Algorithm

1: **Given:** $\mathscr{G} =$ factor graph model,
$\mathbb{S} =$ proper list of separation procedure sets.
2: Construct $G = (V, E, w)$ from $\mathscr{G}$.
3: Initialize the constraint set $\mathscr{C}$ as described in Sec. 5.2.1.
4: **for** $i = 1, \ldots, |\mathbb{S}|$ **do**
5:    **repeat**
6:       Solve $\hat{y} \in \arg\min_{y \in \mathscr{C}} \langle w, y \rangle$,
7:       $\bar{\mathscr{C}} =$ violated constraints found by separation procedures $\mathbb{S}_i$ for $\hat{y}$,
8:       $\mathscr{C} = \mathscr{C} \cup \bar{\mathscr{C}}$,
9:    **until** $\bar{\mathscr{C}}$ is empty.
10: **end for**
11: Compute a labeling $x \in X$ based on $\hat{y}$ (also known as rounding, cf. Sec. 5.3).

---

$a \leq b$ is only added if $a \leq b - \varepsilon$ does not hold. Ignoring this may not only lead to infinite loops for some instances, but may also significantly increase runtime. The parameter $\varepsilon$ should be chosen depending on the precisions of the (I)LP solver. We use $\varepsilon = 10^{-8}$.

### 5.5 Relation between Relaxations of the Multi Way Cut and Marginal Polytope

A major line of research considers relaxations of the marginal polytope to approximately solve problem P1. For the local polytope relaxation [67] of second order models it is known that the local multiway cut relaxation (42) is equivalent if either the labeling is binary [10] or the coupling $w$ is positive [59].

We generalize this results for real valued couplings. This gives a interesting connection between local polytope and multiway cut relaxation and allows us to rank our relaxations. As the proof is very technical, we add the corresponding Thm. 7.1 and its proof in the appendix.

## 6 Experiments

### 6.1 Set-Up, Implementation Details

We implemented the separation procedures and reduction methods described above using C++ and the OpenGM2-library [5] for the factor graph representation, and CPLEX for solving ILPs and LPs in the inner loop of the iteration.

Our multicut approach encompasses a variety of algorithms which differ in the used inequalities, in the separation procedures, and in the order these procedures are applied. The abbreviations for single separation procedures are listed as Tab. 2.

For example, MC-CFB-I-CIF indicates:

- application of the multicut algorithm (MC) based on
- searching for violated facet-defining cycle inequalities (CF) using bounding (B),
- enforcing integer constraints (I), and finally
- searching for facet-defining cycle inequalities violated by the current *Boolean* solution (CIF), based on Breadth-First-Search instead of the Dijkstra algorithm (cf. Sec. 5.2.3).

| | |
|---|---|
| **I** | integer constraints |
| **C** | cycle inequalities separation |
| **CF** | facet-defining cycle inequalities separation |
| **CI** | cycle inequalities separation for ILP |
| **CIF** | facet-defining cycle inequalities separation for ILP |
| **OW** | odd-wheel inequalities separation |
| **T** | terminal inequalities separation |
| **MT** | multi terminal inequalities separation |
| **TI** | terminal inequalities separation for ILP |
| ***B** | bounding for the shortest path search was used |

Table 2: Abbreviations for the separation procedures.

| name | #instances | #nodes | order | type | results | references |
|---|---|---|---|---|---|---|
| synth-potts | 10 | $32^2$ | 2 | US | Tab. 4 | - |
| synth-inclusion | 10 | $32^2$ | 4 | S | Fig. 13, Tab. 10 | - |
| image-seg | 100 | 156-3764 | 2 | US | Fig. 11, Tab. 5 | [4, 36] |
| image-seg$^3$ | 100 | 156-3764 | 3 | US | Fig. 11, Tab. 6 | [4, 36] |
| corr-clust | 715 | 122-651 | 34-651 | US | Fig. 12, Tab. 7 | [40, 36] |
| mod-clust | 6 | 34-115 | 2 | US | Tab. 8 | [14, 38] |
| color-seg | 3 | 424720 | 3 | S | Tab. 9 | [2, 36] |

Table 3: Overview of the datasets used for evaluation.

We report for each dataset results averaged over all its instances:

1. the mean[6] runtime: *runtime*,

2. the mean[6] value (energy) of the integer solution after rounding: *value*,

3. the mean[6] lower bound, given by the solution of the relaxed problem: *bound*,

4. how often the method found an integer solution with an objective value not larger than $10^{-6}$ compared to the overall best method for this instance: *best*[7], and

5. how often the method provided a gap between the objective value of the integer solution and the lower bound, that was smaller than $10^{-6}$: *ver. opt*[7], which we interpret as globally optimal for our instances.

6. if available we also evaluate on an application specific loss, e.g. Variation of Information (VI) [54], Rand Index (RI) [61], and Pixel Accuracy (PA).

In the *unsupervised case*, we compared the proposed methods with our implementation of the Kernighan-Lin (KL) algorithm [39] for the second-order case, as well as with iterative conditional mode (ICM) [9] and Lazy Flipper (LF) [6]. For *planar graphs*, an optimal segmentation with only four labels exists, and methods for the supervised case can be applied.

In the *supervised case*, we compared with TRWS [46], Max Product Linear Programming with no (MPLP) [30] and with cycle-inequalities MPLP-C [64, 65], $\alpha$-expansion [13] and FastPD [48] – using in each case code provided by the respective authors of these papers. Furthermore, we compared to commercial LP- and ILP-solvers in the nodal domain, LBP, TRBP, and $\alpha$-fusion, as provided by OpenGM2.

## 6.2 Unsupervised Segmentation Problems

### 6.2.1 Synthetic Experiments

For numerical evaluation we generate 10 synthetic Potts instances (*synth-potts*). The models have a underlying grid structured (4 neighbors) with $32 \times 32$ variables with 10 labels each. Unary terms are uniformly sample from $[0, 1]$ and the coupling of the pairwise Potts terms are uniformly sampled from $[-1, 1]$. Results averaged over all instances are shown in Tab. 4. MC-T-MT and LP solve both the local polytope relaxation. TRWS and MPLP stops too early or get stuck in local fixed points, which can be seen by the worse bound. When we add cycle constraints, MPLP-C performs slightly better in terms of the lower bound but of cost of significant higher runtime. While the differences in the averaged value are mainly caused by more involved rounding used by MPLP-C, the better bound might be caused by a tighter relaxation. Contrary to our approach MPLP-C also considers violated odd-cycle-constraints on binary partitions of the label spaces, which additionally tightens the polytope for models with more than 2 labels. However, when we add integer constraints after our cut phases the optimum is found for 9 of 10 cases and for the remaining one some additional integer terminal constraints guarantee to find the global optimal solution. MC-T-MT-CFB-I-TI is 100 times faster than the ILP and 10 times faster than MPLP-C. Latter even is not able to find optimal solutions in all cases.

| algorithm | runtime | value | bound | best | ver. opt |
|---|---|---|---|---|---|
| ICM | 0.03 sec | −95.87 | −∞ | 0/10 | 0/10 |
| LF-1 | 0.01 sec | −95.79 | −∞ | 0/10 | 0/10 |
| LP | 2.92 sec | −152.89 | −189.09 | 0/10 | 0/10 |
| MPLP | 5.24 sec | −159.60 | −189.16 | 0/10 | 0/10 |
| TRWS | 0.93 sec | −168.87 | −189.15 | 0/10 | 0/10 |
| MC-T-MT | 1.59 sec | −152.72 | −189.09 | 0/10 | 0/10 |
| MPLP-C | 36.78 sec | −183.90 | −184.14 | 8/10 | 3/10 |
| MC-T-MT-CFB | 2.06 sec | −179.00 | −184.47 | 0/10 | 0/10 |
| ILP | 421.24 sec | **−184.14** | −184.14 | 10/10 | 10/10 |
| MC-T-MT-CFB-I-TI | **4.30** sec | **−184.14** | −184.14 | 10/10 | 10/10 |

Table 4: Synthetic Potts models with 10 labels on a $32 \times 32$ grid

### 6.2.2 Probabilistic Image Segmentation

The probabilistic image segmentation framework was suggested by Andres et al. [4] and belongs to the class of unsupervised image segmentation problems. These problem instances involve $156 - 3764$ superpixels. For all pairs of adjacent superpixels, the likelihood that their common part of the superpixel boundary is part of the segmentation, is learned off-line by a random forest. This results in a Potts model with positive and negative coupling constraints. While the connection to Potts models is not mentioned in [4], they use a similar optimization scheme as in the present work. They introduced a higher-order model as well as a second-order one. They have been made publicly available in [38] and [36], respectively.

**Second-order Case.** As shown in Tab. 5, for this dataset (*image-seg*), we profit from using ILP subproblems. This reduces the mean runtime to less than 3 seconds and is therefore empirically faster than LP-based cutting-plane methods and the heuristic KL-algorithm. ICM and LF perform worse than KL. With increasing search space LF outperforms KL. For a search-depth greater than 1 we make use of the fact that the instances are planar and an optimal solution with four labels exists. The same trick is used to make TRWS applicable. Additionally, we fix the first variable and initialize messages randomly. Even this does not help to prevent TRWS from running into poor local fix-points. In both cases the label reduction is marked by the postfix *L4*.

---

[6] Averaged over all instances of this dataset.

[7] Note, that 0 means that the method has never found the best solution among all methods or has never verified optimal solution. Of course, this does not mean that solutions provided by this method are poor. Performing slightly worse than optimal already returns the value 0.

| algorithm | runtime | value | bound | best | ver. opt | VI | RI |
|---|---|---|---|---|---|---|---|
| KL | 4.96 s | 4608.57 | $-\infty$ | 0.0% | 0.0% | 2.6431 | 0.6401 |
| ICM | 6.03 s | 4705.07 | $-\infty$ | 0.0% | 0.0% | 2.8580 | 0.5954 |
| LF1 | 2.35 s | 4705.01 | $-\infty$ | 0.0% | 0.0% | 2.8583 | 0.5953 |
| LF2-L4 | 0.13 s | 4627.38 | $-\infty$ | 0.0% | 0.0% | 2.9020 | 0.5821 |
| LF3-L4 | 3.16 s | 4581.83 | $-\infty$ | 0.0% | 0.0% | 2.9102 | 0.5873 |
| LF4-L4 | 176.47 s | 4555.73 | $-\infty$ | 0.0% | 0.0% | 2.9164 | 0.5926 |
| TRWS-L4 | 0.84 s | 4889.23 | 4096.53 | 0.0% | 0.0% | 3.2164 | 0.6628 |
| MC-C | 14.02 s | 4447.47 | 4442.34 | 35.0% | 35.0% | 2.5490 | **0.7822** |
| MC-CB | **4.71 s** | 4447.47 | 4442.34 | 35.0% | 35.0% | 2.5490 | **0.7822** |
| MC-CF | 11.35 s | 4447.47 | 4442.34 | 35.0% | 35.0% | 2.5490 | **0.7822** |
| MC-CFB | 5.16 s | 4447.47 | 4442.34 | 35.0% | 35.0% | 2.5490 | **0.7822** |
| MC-C-OW | 14.08 s | 4447.41 | 4442.34 | 35.0% | 35.0% | 2.5489 | **0.7822** |
| MC-CB-OW | **4.81 s** | 4447.41 | 4442.34 | 35.0% | 35.0% | 2.5489 | **0.7822** |
| MC-CF-OW | 11.45 s | 4447.41 | 4442.34 | 35.0% | 35.0% | 2.5490 | **0.7822** |
| MC-CFB-OW | 5.19 s | 4447.41 | 4442.34 | 35.0% | 35.0% | 2.5490 | **0.7822** |
| MC-I-CI | 2.78 s | **4442.64** | **4442.64** | **100.0%** | **100.0%** | 2.5367 | 0.7821 |
| MC-I-CIF | **2.20 s** | **4442.64** | **4442.64** | **100.0%** | **100.0%** | 2.5363 | 0.7821 |
| MC-C-I-CI | 15.00 s | **4442.64** | **4442.64** | **100.0%** | **100.0%** | 2.5365 | 0.7821 |
| MC-CFB-I-CIF | 5.69 s | **4442.64** | **4442.64** | **100.0%** | **100.0%** | 2.5365 | 0.7821 |

Table 5: Second-order probabilistic image segmentation [4, 36]

| algorithm | runtime | value | bound | best | ver. opt | VI | RI |
|---|---|---|---|---|---|---|---|
| ICM | 10.79 s | 6030.49 | $-\infty$ | 0.0% | 0.0% | 2.7089 | 0.5031 |
| LF | 4.17 s | 6030.29 | $-\infty$ | 0.0% | 0.0% | 2.7095 | 0.5033 |
| MC-C | 43.82 s | 6657.32 | 5465.15 | 0.0% | 0.0% | 3.9927 | **0.7755** |
| MC-CB | 42.86 s | 6657.32 | 5465.15 | 0.0% | 0.0% | 3.9927 | **0.7755** |
| MC-CF | 26.68 s | 6658.28 | 5465.15 | 0.0% | 0.0% | 3.9935 | **0.7755** |
| MC-CFB | **25.00 s** | 6658.28 | 5465.15 | 0.0% | 0.0% | 3.9935 | **0.7755** |
| MC-C-OW | 43.71 s | 6657.12 | 5465.29 | 0.0% | 0.0% | 3.9928 | 0.7754 |
| MC-CB-OW | 43.38 s | 6657.12 | 5465.29 | 0.0% | 0.0% | 3.9928 | 0.7754 |
| MC-CF-OW | 27.62 s | 6658.08 | 5465.29 | 0.0% | 0.0% | 3.9936 | 0.7754 |
| MC-CFB-OW | **25.55 s** | 6658.08 | 5465.29 | 0.0% | 0.0% | 3.9936 | 0.7754 |
| MC-I-C | 689.79 s | **5627.52** | **5627.52** | **100.0%** | **100.0%** | 2.6586 | 0.7727 |
| MC-I-CFB | 469.87 s | **5627.52** | **5627.52** | **100.0%** | **100.0%** | 2.6586 | 0.7727 |
| MC-I-CI | 119.64 s | **5627.52** | **5627.52** | **100.0%** | **100.0%** | 2.6586 | 0.7727 |
| MC-I-CIF | **72.81 s** | **5627.52** | **5627.52** | **100.0%** | **100.0%** | 2.6586 | 0.7727 |
| MC-C-I-CI | 125.33 s | **5627.52** | **5627.52** | **100.0%** | **100.0%** | 2.6586 | 0.7727 |
| MC-CFB-I-CIF | 82.00 s | **5627.52** | **5627.52** | **100.0%** | **100.0%** | 2.6586 | 0.7727 |

Table 6: Third-order probabilistic image segmentation [4, 38]

Concerning the multicut approach, odd-wheel constraints only marginally improve the results. LP-based cutting-plane methods find the optimal solution for 35 of 100 instances and are slower than ILP-based methods, too.

**Higher-order Case.** The third-order models (*image-seg*$^3$) from [4] are hard to solve with relaxations, hence rounding becomes more important, cf. Fig. 10. The additional third-order factors favor smooth boundary continuation. Since this sometimes conflicts with local boundary probabilities, the problem becomes more involved.

As shown in Tab. 6, local search methods give better results than relaxed solutions after rounding. Our exact multicut scheme was able to solve all instances to optimality. Notably, one instance was significantly harder than all others and took more than half of the overall runtime for MC-I-C and MC-I-CFB.

Overall, a few instances are significantly harder than others.

Compared to the second order model the Variation of Information (VI) [54] and Rand Index (RI) [61] are worse. The reason might be a to strong regularisation with the boundary continuation which results in segmentation that does not fit with the BSD-ground-truth well.

### 6.2.3 Higher-order Hierarchical Image Segmentation

The hierarchical image segmentation framework was suggested by Kim et al. [40] and also belongs to the class of unsupervised image segmentation problems. Contrary to the work of Andres et al. [4], they learn their model-parameters by a structured support vector machine (S-SVM). Further-

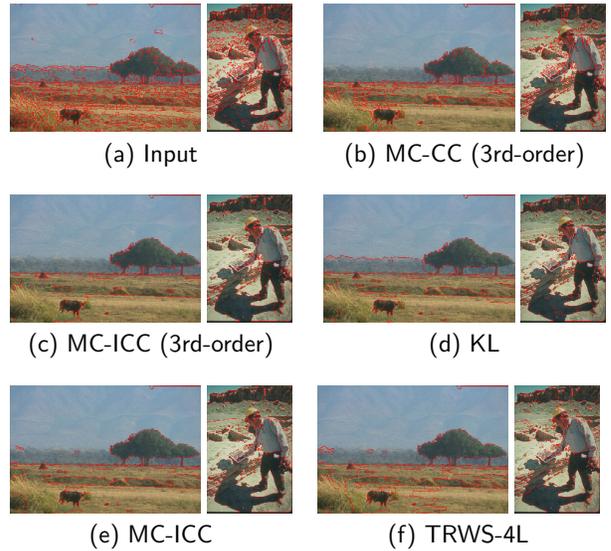

(a) Input    (b) MC-CC (3rd-order)

(c) MC-ICC (3rd-order)    (d) KL

(e) MC-ICC    (f) TRWS-4L

Figure 11: Example for second and third order image segmentation. When the second order model is solved by KL or TRWS (using 4 labels was sufficient) the partitions tend to be over-segmented as compared to the optimal solution obtained by MC-ICC. The 3rd-order model returned segmentations different from those obtained using the second-order model, having higher quality in many but not all regions. This gain of performance could be improved by learning the higher-order model parameters.

| algorithm | runtime | value | bound | best | ver. opt | VI | RI |
|---|---|---|---|---|---|---|---|
| ICM | 1.90 s | $-585.60$ | $-\infty$ | 0.0% | 0.0% | 2.6245 | 0.5154 |
| LF | 1.00 s | $-585.60$ | $-\infty$ | 0.0% | 0.0% | 2.6245 | 0.5154 |
| MC-C | 0.23 s | $-625.97$ | $-628.89$ | 19.9% | 13.7% | 2.0684 | **0.8371** |
| MC-CB | 0.12 s | $-625.97$ | $-628.89$ | 19.9% | 13.7% | 2.0684 | **0.8371** |
| MC-CF | 0.20 s | $-625.97$ | $-628.89$ | 19.9% | 13.7% | 2.0684 | **0.8371** |
| MC-CFB | **0.11 s** | $-625.97$ | $-628.89$ | 19.9% | 13.7% | 2.0684 | **0.8371** |
| MC-C-OW | 0.24 s | $-625.98$ | $-628.89$ | 20.1% | 14.0% | 2.0681 | **0.8371** |
| MC-CB-OW | 0.14 s | $-625.98$ | $-628.89$ | 20.1% | 14.0% | 2.0681 | **0.8371** |
| MC-CF-OW | 0.21 s | $-625.98$ | $-628.89$ | 20.1% | 14.0% | 2.0681 | **0.8371** |
| MC-CFB-OW | **0.13 s** | $-625.98$ | $-628.89$ | 20.1% | 14.0% | 2.0681 | **0.8371** |
| MCR [40] | 0.38 s | $-624.35$ | $-629.03$ | 16.4% | 10.2% | 2.0500 | 0.8357 |
| MC-CI | 1.14 s | $-628.16$ | $-628.16$ | **100.0%** | **100.0%** | 2.0406 | 0.8350 |
| MC-CIF | 1.04 s | $-628.16$ | $-628.16$ | **100.0%** | **100.0%** | 2.0406 | 0.8350 |
| MC-C-CI | 0.85 s | $-628.16$ | $-628.16$ | **100.0%** | **100.0%** | 2.0406 | 0.8350 |
| MC-CFB-CIF | **0.62 s** | $-628.16$ | $-628.16$ | **100.0%** | **100.0%** | 2.0406 | 0.8350 |

Table 7: Higher-order hierarchical image segmentation [40, 36].

more, higher-order Potts terms force selected regions to belong to the same cluster. The 715 instances of this dataset (*corr-clust*), published as part of [36], contain factors of order up to a few hundred and 122–651 variables.

The results are summarized as Tab. 7. Surprisingly, our LP-based methods perform better than the original algorithm used in [40], even though the algorithms are identical. Maybe this was caused by the different LP solver they used, or by some floating-point problems inside their separation procedure. The use of odd-wheel constraints marginally improves the results. Best results are obtained by using integer cutting-planes after having solved the LP. The use of the bounding as part of the post-processing reduces runtime by a factor of 2. The differences to only using facet-defining constraints are negligible.

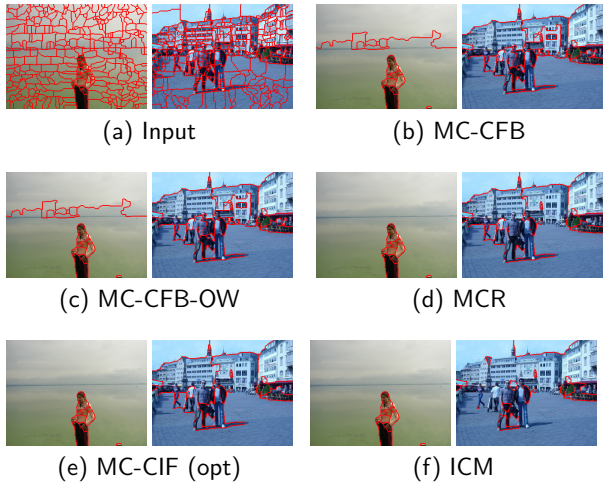

(a) Input (b) MC-CFB
(c) MC-CFB-OW (d) MCR
(e) MC-CIF (opt) (f) ICM

Figure 12: Two instances of the hierarchical correlation-clustering problem. The first image is easier and ICM gave competitive results relative to multicut methods. This is no longer true for more complicated images like the second one where ICM clearly is inferior. MC-CFB and MC-CFB-OW make use of slightly different rounding procedures that result in finer segmentations. MC-CIF computes the segmentation with the optimal energy, which differs slightly from those of MCR but improves the VI, cf. Tab. 7. This small improvement is remarkable, since the models are trained for MCR. So the models are biased towards this method and hence it is not surprising that it performs well.

#### 6.2.4 Modularity Clustering

We also considered a clustering problem from outside the field of computer vision, which contrary to the previous models considered so far, involves a fully connected graph. Modularity clustering [14] means the problem of clustering an undirected unweighted graph into "meaningful" subsets, which amounts to optimization problems related to fully connected Potts model. For our experiments, we used the datasets[8] *dolphins*, *football*, *karate*, and *lesmis* (*mod-clust*), with 62, 115, 34, and 77 data-points, respectively.

As shown in Tab. 8, for modularity clustering, the use of facet-defining inequalities as well as odd-wheel constraints significantly improves the results. We attribute this to the high connectivity of the graph. In such dense graphs more likely violated odd-wheel inequalities exist. Likewise, more *non*-facet-defining cycle inequalities exist as well, and adding those only blows up the system of inequalities.

As observed by Nowozin and Jegelka [58], odd-wheel inequalities usually tighten sufficiently the polytope. Furthermore, we observed for this dataset, as in [58], numerical problems if the allowed feasibility and optimality tolerances were set to large. However, the experiments showed that our proposed integer cycle inequalities perform better than odd-wheel separation, especially if we start from the LP-relaxation with cycle inequalities, cf. Tab. 8.

---

[8] http://www-personal.umich.edu/~mejn/netdata/

| algorithm | runtime | value | bound | best | ver. opt |
|---|---|---|---|---|---|
| KL | **0.01 s** | −0.5251 | −∞ | 2/4 | 0/4 |
| ICM | 0.12 s | 0.0000 | −∞ | 0/4 | 0/4 |
| LF | 0.05 s | 0.0000 | −∞ | 0/4 | 0/4 |
| MC-C | 47.99 s | −0.5204 | −0.5294 | 1/4 | 1/4 |
| MC-CB | 48.33 s | −0.5204 | −0.5294 | 1/4 | 1/4 |
| MC-CF | 1.02 s | −0.5204 | −0.5294 | 1/4 | 1/4 |
| MC-CFB | 0.91 s | −0.5204 | −0.5294 | 1/4 | 1/4 |
| MC-C-OW | 72.05 s | −0.5282 | −0.5282 | 4/4 | 4/4 |
| MC-CB-OW | 72.42 s | −0.5282 | −0.5282 | 4/4 | 4/4 |
| MC-CF-OW | 12.26 s | −0.5282 | −0.5282 | 4/4 | 4/4 |
| MC-CFB-OW | 11.60 s | −0.5282 | −0.5282 | 4/4 | 4/4 |
| MC-I-C | 152.20 s | −0.5282 | −0.5282 | 4/4 | 4/4 |
| MC-I-CI | 14.57 s | −0.5282 | −0.5282 | 4/4 | 4/4 |
| MC-I-CIF | 6.31 s | −0.5282 | −0.5282 | 4/4 | 4/4 |
| MC-I-CFDB | 6.56 s | −0.5282 | −0.5282 | 4/4 | 4/4 |
| MC-C-I-CI | 58.24 s | −0.5282 | −0.5282 | 4/4 | 4/4 |
| MC-CFB-I-CIF | **1.31 s** | −0.5282 | −0.5282 | 4/4 | 4/4 |

Table 8: Modularity clustering [14, 38]

| algorithm | runtime | value | bound | best | ver. opt |
|---|---|---|---|---|---|
| FastPD | **0.45 s** | 308 472 275.0 | −∞ | 2/3 | 0/3 |
| FastPD* | 1.62 s | 308 472 274.7 | −∞ | 2/3 | 0/3 |
| α-Exp | 6.42 s | 308 472 275.6 | −∞ | 2/3 | 0/3 |
| α-Exp* | 1.72 s | **308 472 274.3** | −∞ | 3/3 | 0/3 |
| MC-T-MT | 115.14 s | 308 472 274.3 | 308 472 274.3 | 3/3 | 3/3 |
| MC*-T-MT | **1.76 s** | 308 472 274.3 | 308 472 274.3 | 3/3 | 3/3 |
| LP | † | † | † | † | † |
| LP* | 2.17 s | 308 472 274.3 | 308 472 274.3 | 3/3 | 3/3 |
| TRWS | 150.47 s | 308 472 310.6 | 308 472 270.4 | 2/3 | 1/3 |
| TRWS* | 3.90 s | 308 472 274.3 | 308 472 274.3 | 2/3 | 2/3 |
| MC-T-MT-I-T | 149.43 s | 308 472 274.3 | 308 472 274.3 | 3/3 | 3/3 |
| MC*-T-MT-I-T | **1.86 s** | 308 472 274.3 | 308 472 274.3 | 3/3 | 3/3 |
| ILP | † | † | † | † | † |
| ILP* | 1.91 s | **308 472 274.3** | 308 472 274.3 | 3/3 | 3/3 |

Table 9: Supervised image segmentation [2, 36]

### 6.3 Supervised Segmentation Problems

#### 6.3.1 Supervised Image Segmentation

An elementary approach to supervised image segmentation, or image labeling, is to apply locally a statistical classifier, trained offline beforehand, to raw image data or to locally extracted image features. This is complemented by a non-local prior term, the most common form of which favours short boundaries of the segments partitioning the image domain. Such terms can be approximated by pairwise Potts terms [11] and lead to an energy function of the form

$$\sum_{f \in \mathcal{F}_1} -\log(p_{ne(f)}(x_{ne(f)}|I)) + \sum_{f \in \mathcal{F}_2} \beta \, \mathbb{I}(x_{ne(f)_1} \neq x_{ne(f)_2}). \tag{40}$$

As recently shown by Kappes et al. [37], such models can be evaluated globally optimal and very fast by first determining partial optimality, leading to a reduced inference problem in terms of remaining unlabelled connected image components, followed by solving each of these smaller problems independently.

We use *"∗"* to mark when these preprocessing steps were applied and *"†"* to mark whenever the memory requirement exceeded 12 GB.

As dataset (*color-seg*) we used the color segmentation instances of Alahari et al. [2]. The results are summarized as Tab. 9.

While standard (I)LP solvers often suffer from their large memory requirements, the multicut approach outperformed all other approaches. Since for all instances the local polytope relaxation returned optimal integer solutions, MC-T-MT could solve them in polynomial time. When we resorted to the model reduction *, the subproblems became small for these problem instances, and (I)LP solvers could be conveniently applied. Our multicut approach then was only marginally faster. Despite global optimality, however, the runtime was comparable to algorithms for approximate inference that do not guarantee global optimality.

### 6.3.2 Higher-Order Supervised Image Segmentation with Inclusion Prior

We studied image segmentation with junction regularisation as problem instances that benefit from the application of higher-order generalized Potts functions.

Rather than merely penalizing the boundary length of segments, this approach aims at improving segmentation results by additionally penalizing points where the boundaries of three or more segments meet:

$$\varphi^I(x_1,x_2,x_3,x_4) = \begin{cases} \lambda, & \text{if } |\{x_1,x_2,x_3,x_4\}| > 2, \\ 0, & \text{else.} \end{cases} \quad (41)$$

The overall cost for labeling then is given by

$$\sum_{f \in \mathscr{F}_1} \varphi^1_f(x_{ne(f)}) + \sum_{f \in \mathscr{F}_2} \varphi^2_f(x_{ne(f)}) + \sum_{f \in \mathscr{F}_4} \varphi^I(x_{ne(f)}),$$

where $\varphi^1$ denotes the $L_1$-norm of the difference between intensity of a pixel and a pixel-label, $\varphi^2$ the same second-order terms as in the pairwise case, and $\mathscr{F}_4$ the set of all factors over four pixels that build a cycle in the image grid.

Setting $\lambda$ to 0 yields standard second-order model with boundary length regularization, whereas setting $\lambda \to \infty$ yields a model that enforces segments to be surrounded by one single segment, without fixing the topology of the inclusion as done by Delong et al. [20]. Contrary to [20] our model "learns" the geometric interaction *"contain"* [20] locally on-line and allows furthermore to use different and unknown "containing rules" in different regions instead a single global one.

Fig. 1(b) illustrates this property of the model. The standard second-order approach, cf. Fig. 1(b), middle, produces many small artefacts inside "U" and "C" and opens the surrounding segment right of "C". Invoking the fourth-order regularizer, cf. Fig. 1(b), bottom, eliminates many of these artefacts and results in a significantly better segmentation.

The results of an empirical evaluation for 10 synthetic $32 \times 32$ images (*synth-inclusion*) are summarized as Tab. 10. Fig. 13 show exemplary segmentation results. Additionally, we give the relative number of correct labeled pixels as pixel accuracy (PA). Even a labeling with high energy can have a high PA, as happens for MPLP. This is caused by some variables, with wrong labels that causes a high energy but count marginal for the PA.

Approximate inference methods performed quite good, but among those only LBP-LF2 (Lazy Flipper initialed with the solution of LBP) was able to provide nearly optimal results. While the multicut approach is on par when relaxations were considered, it became quite slow compared to a ILP applied to labeling in the nodal domain, when a globally optimal solution was enforced.

We believe there are two major reasons: First, the relaxation "prefers" less integral solutions due to the higher-order terms and therefore becomes harder to solve for LP-based methods. Second, we observe that CPLEX solves the ILP mainly by branching and probing in order to avoid solving LPs. This is also the reason why ILP is faster than LP.

| algorithm | runtime | value | bound | best | ver. opt | PA |
|---|---|---|---|---|---|---|
| ogm-ICM | 0.03 sec | 1556.20 | $-\infty$ | 0/10 | 0/10 | 0.6206 |
| ogm-LF-1 | 0.04 sec | 1556.20 | $-\infty$ | 0/10 | 0/10 | 0.6206 |
| LBP-LF-2 | 12.20 sec | 1400.62 | $-\infty$ | 8/10 | 0/10 | 0.9495 |
| $\alpha$-Fusion | 0.07 sec | 1587.13 | $-\infty$ | 0/10 | 0/10 | 0.6771 |
| ogm-LBP | 12.28 sec | 1800.67 | $-\infty$ | 3/10 | 0/10 | 0.9495 |
| ogm-TRBP | 13.93 sec | 2000.67 | $-\infty$ | 2/10 | 0/10 | 0.9491 |
| MC-T-MT | 18.55 sec | 1739.29 | 1399.49 | 1/10 | 0/10 | 0.3001 |
| LP | 25.04 sec | 3900.59 | 1400.33 | 1/10 | 1/10 | 0.9484 |
| MPLP | 10.08 sec | 4000.44 | 1400.30 | 1/10 | 1/10 | 0.9479 |
| MPLP-C | 4741.42 sec | 4000.41 | 1400.35 | 1/10 | 1/10 | 0.9471 |
| ILP | **7.33** sec | **1400.57** | 1400.57 | 10/10 | 10/10 | 0.9496 |
| MC-T-MT-I-T | 66.58 sec | **1400.57** | 1400.57 | 10/10 | 10/10 | 0.9496 |

Table 10: Supervised image segmentation with inclusion priors.

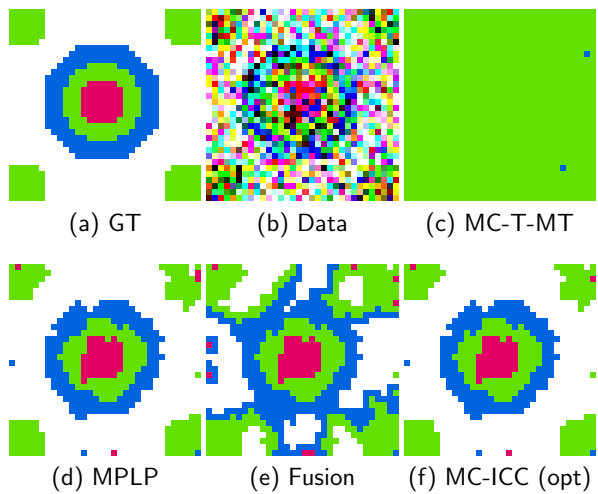

(a) GT    (b) Data    (c) MC-T-MT

(d) MPLP    (e) Fusion    (f) MC-ICC (opt)

Figure 13: Example instance for the inclusion problem. While MC-T-MT and $\alpha$-Fusion both found solutions that respect the inclusion prior, they do not fit the data term well. MPLP found a good solution but does not respect the inclusion in the right upper corner. MC-ICC found the optimal solution of the problem which visually fulfils inclusion and gave good results in view of the noise level.

While an in-depth study of such aspects is beyond the scope of the present paper, our findings indicate ways to further improve the multicut approach in such advanced settings.

## 7 Conclusion

We presented an approach based on multicuts to solve a broad range of supervised and unsupervised segmentation problems to optimality in reasonable runtime. We showed, in particular, how to extend the approach higher-order models based on a class of label invariant functions that generalize Potts functions in a natural way. Such models enable to model higher-order interactions and topological priors concisely by taking its symmetries into account.

We devised several dedicated separation procedures and demonstrated a corresponding significant impact on runtime. A systematic comparison of different cutting-plane procedures for computer vision applications enabled us to improve runtimes for all models compared to the state of the art. A discussion of polynomially solvable relaxations of the unsupervised segmentation problems complemented our study, together with advanced rounding schemes.

**Acknowledgments.** This work has been supported by the German Research Foundation (DFG) within the program "Spatio-/Temporal Graphical Models and Applications in Image Analysis", grant GRK 1653.

# Appendix

**Theorem 7.1.** *For second-order Potts models with arbitrary coupling constraints $w \in \mathbb{R}^N$ the optima of the LP over the local multiway cut relaxation in (42)*

$$\min_{y \in [0,1]^N} \langle w, y \rangle \tag{42}$$

$$\text{s.t.} \sum_{t \in T} y_{tv} = |T| - 1, \text{ if } T \neq \emptyset, \forall v \in V \setminus T \text{ c.f. (26)}$$

$$y_{tt'} = 1, \qquad \forall t, t' \in T, t \neq t' \text{ c.f. (27)}$$

$$y_{tu} + y_{tv} \geq y_{uv}, \qquad \forall uv \in E, t \in T \text{ c.f. (30)}$$

$$y_{tu} + y_{uv} \geq y_{tv}, \qquad \forall uv \in E, t \in T \text{ c.f. (31)}$$

$$y_{tv} + y_{uv} \geq y_{tu}, \qquad \forall uv \in E, t \in T \text{ c.f. (32)}$$

$$y_{uv} \geq \sum_{t \in S} (y_{tu} - y_{tv}), \quad \forall uv \in E, S \subseteq T \text{ c.f. (33)}$$

*is equivalent to the optima of LP over the local polytope relaxation (43)*

$$\min_{\mu \in [0,1]^N} \langle \theta, \mu \rangle. \tag{43}$$

$$\text{s.t.} \sum_{x_i \in X_i} \mu_{i;x_i} = 1 \qquad \forall i \in V$$

$$\sum_{x_i \in X_i} \mu_{ij;x_i x_j} = \mu_{j;x_j} \qquad \forall f \in \mathscr{F}_2, \{i,j\} = ne(f)$$

$$\sum_{x_j \in X_j} \mu_{ij;x_i x_j} = \mu_{i;x_i} \qquad \forall f \in \mathscr{F}_2, \{i,j\} = ne(f).$$

*Proof.* By construction of the multiway cut problem we have

$$\mu_{u;i} = 1 - y_{t_i, u} \qquad \forall u \in \mathscr{V}, i \in X_u, \tag{44}$$

where $t_i \in T$ denotes the terminal node belonging to label $i \in X_u$.

Instead of optimizing over $\mu_{u;i}$ for all $u \in V$ and $x_u \in X_u$ we show that for any fixed $\mu_{u;i} \forall u \in \mathscr{V}, i \in X_u$ problem (42) and (43) have the same minima. For fixed unary variables the problem splits in several small terms, which can be treated independently. What is left to show is that for any

$$\theta_{uv} = \begin{cases} w_{uv} & \text{if } u \neq v \\ 0 & \text{if } u = v \end{cases}$$

where $w_{uv} \in \mathbb{R}$ the following equality holds:

$$\min_{\mu_{uv}} \langle \theta_{uv}, \mu_{uv} \rangle \tag{45}$$

$$\text{s.t.} \sum_i \mu_{uv;ij} = \mu_{v,j} \forall j \in X_v,$$

$$\sum_j \mu_{uv;ij} = \mu_{u,i} \forall i \in X_u, \mu_{uv;ij} \geq 0$$

$$= \min_{y_{uv} \in [0,1]} w \cdot y_{uv} \tag{46}$$

$$\text{s.t. } (26), (27), (30) - (32), (33)$$

**Case 1 ($w_{uv} \geq 0$):**
This has been shown by Osokin et al. [59].

**Case 2 ($w_{uv} < 0$):**
The variable $y_{uv}$ is only upper bounded by $[0,1]$ and (30). Furthermore $y$ can be substituted by $\mu$ with (44). Hence the the optimal value of (46) $y_{uv}$ is

$$(46) \stackrel{(30)}{=} w_{uv} \cdot \min\{1, \min_i \{(y_{u,t_i} + y_{v,t_i})\}\} \tag{47}$$

$$\stackrel{(44)}{=} w_{uv} \cdot \min\{1, \min_i \{2 - (\mu_{u,i} + \mu_{v,i})\}\} \tag{48}$$

**Inequality (i):** $(45) \geq (46)$
For this proof we will use the following observation:

**Lemma 7.1.** *Let us consider problem* (45). *If $\mu_{u,i} + \mu_{v,i} > 1$ then $\mu_{uv,ii} \geq \mu_{u,i} + \mu_{v,i} - 1$.*

*Proof.* Let $\mu_{u,i} + \mu_{v,i} > 1$ and w.l.o.g. $\mu_{u,i} > \mu_{v,i}$. Then we have $\mu_{u,i} > 0.5$ and $\mu_{u,i} > 1 - \mu_{v,i}$. Latter causes that we have to put some mass of $\mu_{u,i}$ on the main diagonal of $Q$ since we can assign to the non-diagonal elements only $1 - \mu_{v,i}$. At least the difference $\mu_{u,i} - (1 - \mu_{v,i})$ has to be put on the main diagonal entry $\mu_{uv,ii}$. □

Using Lemma 7.1 we obtain the inequality:

$$(45) = w_{uv} \cdot (1 - \sum_i \mu_{uv,ii}) \text{ s.t. } (45) \tag{49}$$

$$\stackrel{Lemma 7.1}{\geq} w_{uv} \cdot (1 - \sum_i \max\{0, \mu_{u,i} + \mu_{v,i} - 1\}) \tag{50}$$

$$= w_{uv} \cdot (1 - \max_i \max\{0, \mu_{u,i} + \mu_{v,i} - 1\}) \tag{51}$$

$$= w_{uv} \cdot (1 + \min_i \min\{0, -(\mu_{u,i} + \mu_{v,i}) + 1\}) \tag{52}$$

$$= w_{uv} \cdot (\min_i \min\{1, 2 - (\mu_{u,i} + \mu_{v,i})\}) \tag{53}$$

$$= (46) \tag{54}$$

**Inequality (ii):** $(45) \leq (46)$
We will use the following observations next:
Let $i^* = \arg\max_i (\mu_{u,i} + \mu_{v,i})$
(a) If $\max_i (\mu_{u,i} + \mu_{v,i}) < 1$, then $\forall i : (\mu_{u,i} + \mu_{v,i}) < 1$.
(b) If $\max_i (\mu_{u,i} + \mu_{v,i}) > 1$, then $\forall i \neq i^* : (\mu_{u,i} + \mu_{v,i}) < 1$.

**Lemma 7.2.** *Problem* (45) *with the additional constraints*

$$\mu_{uv,ii} = \max\{0, \mu_{u,i} + \mu_{v,i} - 1\} \qquad \forall i \in X_v \tag{55}$$

*has still a feasible solution.*

*Proof.* We split the proof in to cases:
**Case 1:** $\max_i(\mu_{u,i} + \mu_{v,i}) > 1$
Let $i^* = \arg\max_i \mu_{u,i} + \mu_{v,i}$, then

$$\mu_{uv;ij} = \begin{cases} \max_i \mu_{u,i} + \mu_{v,i} - 1 & \text{if } i = i^* \text{ and } j = i^* \\ \mu_{u;i} & \text{if } i \neq i^* \text{ and } j = i^* \\ \mu_{v;j} & \text{if } i = i^* \text{ and } j \neq i^* \\ 0 & \text{else} \end{cases}$$

is a feasible solution.

**Case 2:** $\max_i \mu_{u,i} + \mu_{v,i} \leq 1$
Let us start with any feasible solution $\mu^0$ of (45) and define a sequence of transitions $\mu^n \to \mu^{n+1}$, which will end in a $\mu^m$ that fulfill the additionally constraints ($\forall i : \mu_{uv;ii} = 0$), too.

**Initial point ($\mu^o$):** As shown in [59] a feasible point $\mu^0_{uv,ii}$ of (45) with $\mu^0_{uv,ii} = \min\{\mu_{u;i}, \mu_{v;j}\}$ exists.

**Transition ($\mu^n \to \mu^{n+1}$):** For any tuple $(a,b,a',b')$ and $\delta \leq \min\{mu^n_{uv;ab}, mu^n_{uv;a'b'}\}$ the transition

$$\mu^{n+1}_{uv;ij} = \begin{cases} \mu^n_{uv;ij} - \delta & \text{if } i = a \wedge j = b \text{ or } i = a' \wedge j = b' \\ \mu^n_{uv;ij} + \delta & \text{if } i = a \wedge j = b' \text{ or } i = a' \wedge j = b \\ \mu^n_{uv;ij} & \text{else} \end{cases}$$

stays in the feasible set, because all values remain non-negative and row- and column-sums does not change.

**Sequence of transitions:** When $\mu^n$ has a non-zero diagonal element $\mu^n_{uv;ii}$ then there exist a pair $(i', j')$ with $i \neq i'$, $j \neq j'$ and $\mu^n_{uv;i'j'} > 0$, because

$$\sum_{(a,b), a=i \text{ or } b=j} \mu^n_{uv;ab} = \mu^n_{uv;ii} + (\mu_{u,i} - \mu^n_{uv;ii}) + (\mu_{v,j} - \mu^n_{uv;ii}) \tag{56}$$

$$= \mu_{u,i} + \mu_{v,j} - \mu^n_{uv;ii} \geq 1 - \mu^n_{uv;ii} \tag{57}$$

After the transition on $(i, j, i', j')$ with maximal $\delta$, either $\mu^{n+1}_{uv;ii} = 0$ or $\mu^{n+1}_{uv;i'i'} = 0$. So, each non-zero diagonal element $\mu^n_{uv;ii}$, can be made zero by a finite number of transitions. We will end up with a point $\mu^m$ that fulfill (45) and have zero diagonal element $\mu^n_{uv;ii}$.

□

Using Lemma 7.2 we obtain the inequality:

$$(45) = w_{uv} \cdot \left(1 - \sum_i \mu_{uv,ii}\right) \text{ s.t. (45)} \tag{58}$$

$$\stackrel{Lemma 2}{\leq} w_{uv} \cdot (1 - \max\{0, \max_i(\mu_{u,i} + \mu_{v,i} - 1)\}) \tag{59}$$

$$= w_{uv} \cdot \min\{1, \min_i\{2 - (\mu_{u,i} + \mu_{v,i})\}\} \tag{60}$$

$$= (46) \tag{61}$$

Inequality (i) and inequality (ii) together imply that $(45) = (46)$

□